\documentclass{bmvc2k}

\usepackage{graphicx}
\usepackage{amsmath}
\usepackage{amssymb}
\usepackage{booktabs}
\usepackage{multirow}
\usepackage{enumitem}
\usepackage{verbatim}
\usepackage{comment}
\usepackage{bm}
\usepackage{adjustbox}
\usepackage{array}
\usepackage{pifont}
\usepackage{stmaryrd}
\usepackage{flushend}
\usepackage{makecell}
\usepackage{float}
\usepackage{xfrac}
\usepackage{rotating}
\usepackage{colortbl}
\usepackage[normalem]{ulem}

\usepackage{pifont}
\usepackage{setspace}
\usepackage{cuted}
\usepackage{bm}

\usepackage[dvipsnames]{xcolor}

\newcommand\Xuanlong{\textcolor{black}}
\newcommand\Gianni{\textcolor{black}}
\newcommand\david{\textcolor{black}}
\newcommand{\comEmi}[1]{\textcolor{black}{{#1}}}

\newcommand\AnyV{\textcolor{black}}

\newcommand{\Emi}[1]{#1}

\newcommand{\ab}[1]{\textcolor{black}{#1}}

\newcommand\blfootnote[1]{%
\begingroup
\renewcommand\thefootnote{}\footnote{#1}%
\addtocounter{footnote}{-1}%
\endgroup
}

\title{MUAD: Multiple Uncertainties for Autonomous Driving, a benchmark for multiple uncertainty types and tasks}

\addauthor{Gianni Franchi}{gianni.franchi@ensta-paris.fr}{1$\dagger$}
\addauthor{Xuanlong Yu}{xuanlong.yu@universite-paris-saclay.fr}{1,2$\dagger$}
\addauthor{Andrei Bursuc}{andrei.bursuc@valeo.com}{3}
\addauthor{Angel Tena}{angel.tena@anyverse.ai}{4}
\addauthor{R\'{e}mi Kazmierczak}{remi.kazmierczak@ensta-paris.fr}{1}
\addauthor{S\'{e}verine~Dubuisson}{severine.dubuisson@lis-lab.fr}{5}
\addauthor{Emanuel Aldea}{emanuel.aldea@universite-paris-saclay.fr}{2}
\addauthor{David Filliat}{david.filliat@ensta-paris.fr}{1}
\addinstitution{U2IS, ENSTA Paris, IP Paris}
\addinstitution{SATIE, Paris-Saclay University}
\addinstitution{valeo.ai}
\addinstitution{Anyverse}
\addinstitution{Aix Marseille University}

\begin{small}
\runninghead{MUAD}{Dataset with Multiple Uncertainties for Autonomous Driving}
\end{small}

\def\eg{\emph{e.g}\bmvaOneDot}
\def\Eg{\emph{E.g}\bmvaOneDot}
\def\etal{\emph{et al}\bmvaOneDot}

\definecolor{dgreen}{rgb}{0.0, 0.5, 0.0}
\definecolor{better}{rgb}{0.19, 0.55, 0.91}
\definecolor{worse}{rgb}{0.82, 0.1, 0.26}

\newcommand{\parag}[1]{\smallskip\noindent\textbf{#1}~~}

\newcommand{\second}{\cellcolor{blue!10}}
\newcommand{\first}{\cellcolor{blue!30}}
\begin{document}

\maketitle
\blfootnote{$\dagger$ Equal contribution.}

\begin{figure}[htbp]
                 \centering
                    \subfigure[No perturbation]{ \begin{minipage}{0.23\textwidth} \includegraphics[width=1.05\linewidth]{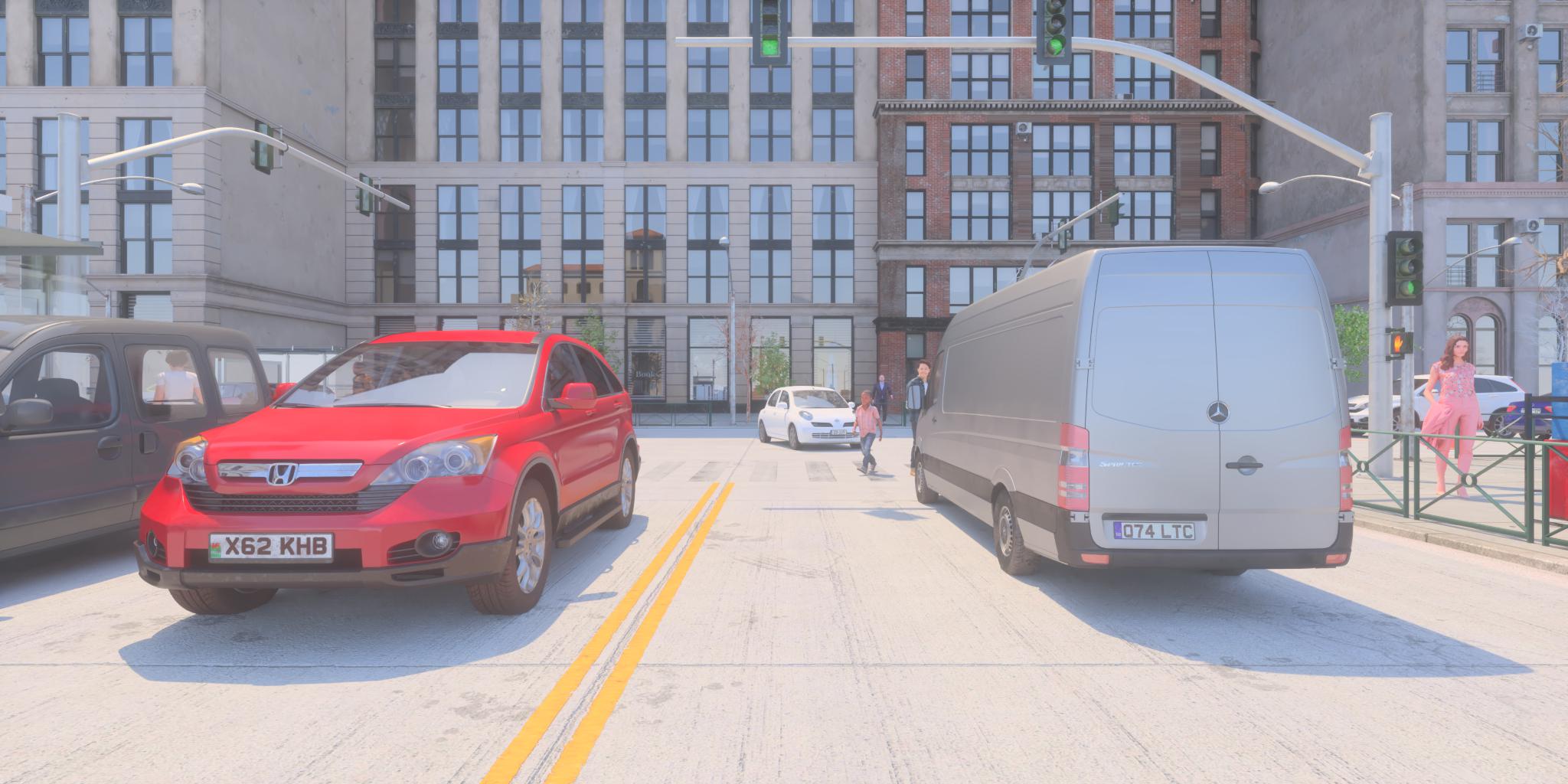}
                      \label{normal_img}
                      \end{minipage}
                     }
                    \subfigure[With OOD instances]{ \begin{minipage}{0.23\textwidth} \includegraphics[width=1.05\linewidth]{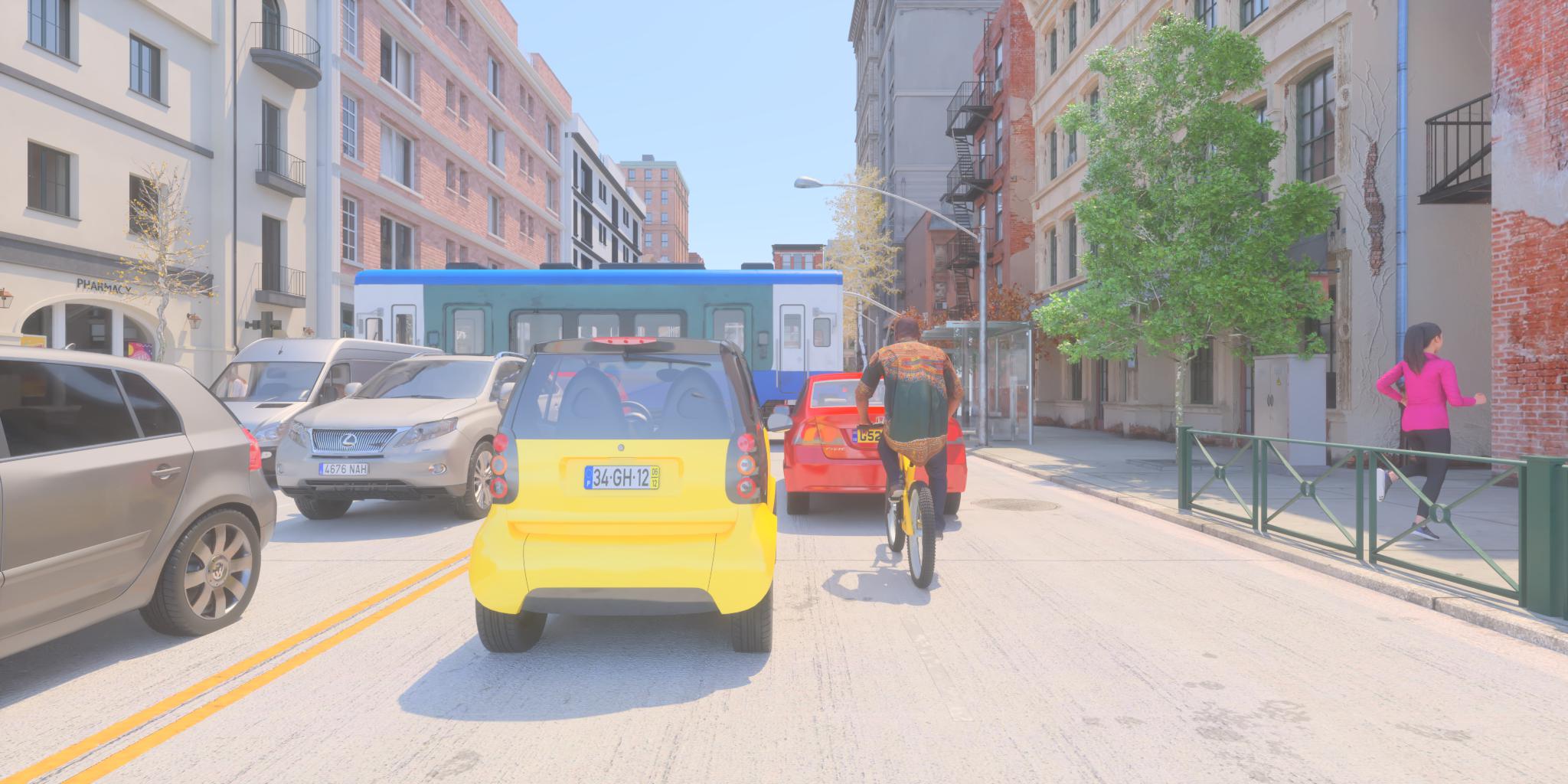}
                      \label{OOD_img}
                      \end{minipage}
                      }
                    \subfigure[Level 1 perturbation]{ \begin{minipage}{0.23\textwidth} \includegraphics[width=1.05\linewidth]{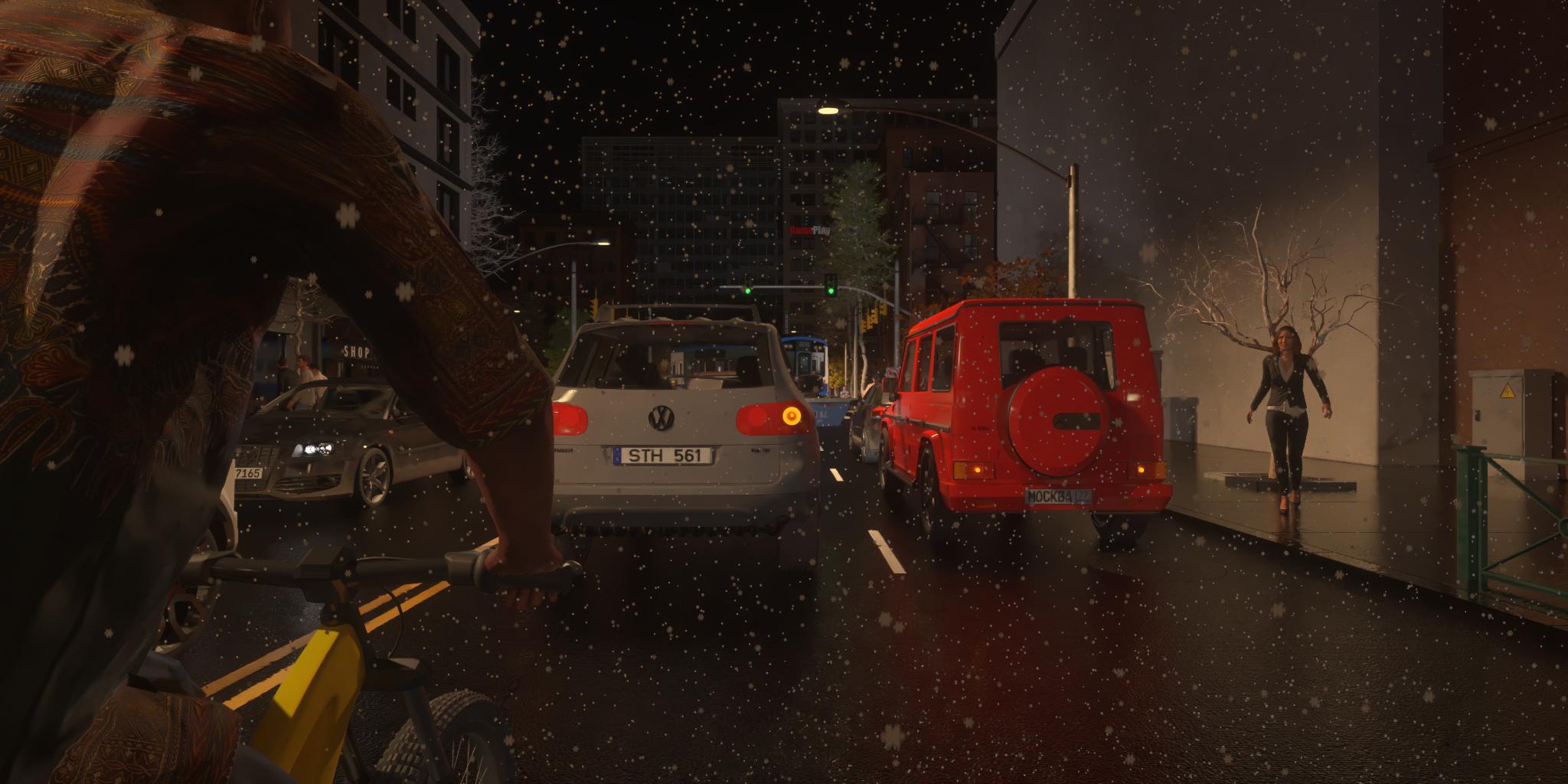}
                      \label{OOD_img}
                      \end{minipage}
                      }
                     \subfigure[Level 2 perturbation]{ \begin{minipage}{0.23\textwidth} \includegraphics[width=1.05\linewidth]{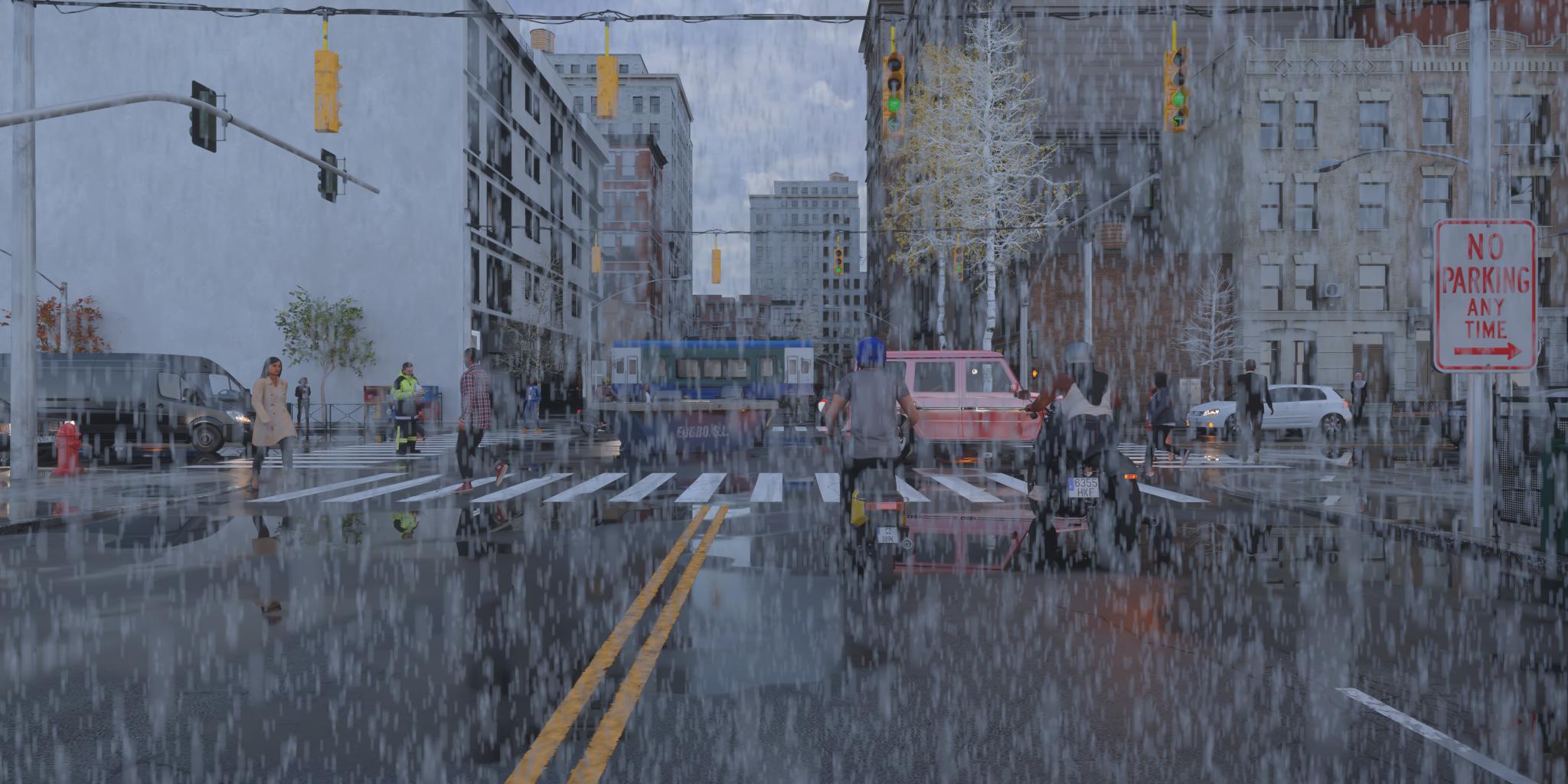}
                      \label{OOD_img}
                      \end{minipage}
                      }
                      \centering
                      \caption{{\small \textbf{Snapshots from the MUAD dataset} showing different types of adverse conditions and events to evaluate perception models (OOD : Out Of Distribution, i.e., not seen during training).}}
                    \vspace{-4mm}
                      \label{fig:teaser}
\end{figure}

\begin{abstract}
Predictive uncertainty estimation is essential for 
\ab{safe deployment of} Deep Neural Networks in real-world autonomous systems. However, disentangling the different types and sources of uncertainty is non trivial 
\ab{for} most datasets, especially since there is no ground truth for uncertainty. 
\ab{In addition, while adverse weather conditions of varying intensities can disrupt neural network predictions, they are usually under-represented in both training and test sets in public datasets.}
\ab{We attempt to mitigate these setbacks and} introduce the MUAD dataset (Multiple Uncertainties for Autonomous Driving), consisting of 10,413 realistic synthetic images with diverse adverse weather conditions (night, fog, rain, snow), out-of-distribution objects and annotations for semantic segmentation, depth estimation, object and instance detection. MUAD allows to better assess the impact of different sources of uncertainty on model performance. 
\ab{We conduct a thorough experimental study of this} impact on several baseline Deep Neural Networks 
\ab{across multiple}
tasks, and \ab{release} our dataset to allow researchers to benchmark their algorithm methodically in adverse conditions.
\Xuanlong{More visualizations and the download link for MUAD are available \comEmi{at} \href{https://muad-dataset.github.io/}{https://muad-dataset.github.io/}.} 
\end{abstract}

\section{Introduction}

In recent years, Deep Neural Networks (DNNs) have achieved remarkable results in various computer vision tasks \cite{krizhevsky2012imagenet,ren2015faster,chen2017deeplab}. This has turned DNNs into an essential tool for effective automatic perception. Although DNNs achieve outstanding performance across benchmarks and tasks, there are still a few major bottlenecks to solve before a widespread deployment. One of the most frequent and known criticisms of DNNs is related to \comEmi{their} lack of reliability \ab{under varying levels of shifts in the data distribution}\comEmi{, and it became} crucial to \comEmi{address this limitation}. To achieve this, we focus on studying the uncertainties of the DNN predictions for computer vision tasks.
The predictive uncertainty of a DNN stems from two main types of uncertainty~\cite{hora1996aleatory}: \ab{\emph{aleatoric} and \emph{epistemic}}.
The former is related to randomness \ab{of the world and of the sensing system}, typically
\ab{instantiated as noise in the data.}
The latter concerns finite size training datasets. The epistemic uncertainty captures the uncertainty in the DNN parameters and their lack of knowledge on the model that generated the training data.
\ab{In spite of their simple and intuitive definitions, the sources of uncertainty are notoriously hard to separate in most datasets, where data are typically curated and various outlier or noisy samples are removed before annotation.}

\ab{For autonomous driving,} uncertainty estimation and reliability are essential for \ab{safely} deploying 
DNNs \ab{in real-world conditions.}
\ab{Here, DNNs are expected not only to reach high predictive performance and real-time inference speed, but also to deal effectively with the two types of uncertainty under various forms (noise, distribution shift, out-of-distribution samples, sensor degradation, etc.).}
\ab{In the last years, numerous works have moved the needle towards more reliable predictive uncertainty for DNNs} \cite{blundell2015weight, gal2016dropout,lakshminarayanan2017simple,kendall2017uncertainties,wen2020batchensemble,franchi2020encoding,mukhoti2020calibrating,besnier2021triggering,mukhoti2021deterministicDDU}. 
However, evaluating such methods is not obvious as there is no ground truth for uncertainty \ab{and the different sources of uncertainty are conflated due to prior data  curation}.

We introduce a new dataset to study 
uncertainty estimation methods 
\ab{for perception in autonomous vehicles.}
While most datasets 
aim to improve the predictive performance of DNNs \cite{geiger2012we,cordts2016cityscapes,neuhold2017mapillary,yu2020bdd100k}, \ab{only recently datasets addressed the robustness of DNNs under unseen weather conditions~\cite{sakaridis2020map,dai2020curriculum,sakaridis2021acdc} or objects~\cite{hendrycks2019anomalyseg,blum2019fishyscapes,chan2021segmentmeifyoucan}.
} 
However, these datasets are either limited to only one task, typically semantic segmentation, or only focus on a single type of uncertainty, or are not being precise enough in the different levels of uncertainties. We address these limitations in our dataset that allows to quantify all levels of uncertainty in the same conditions. Our dataset, MUAD (Multiple Uncertainties for Autonomous Driving) is composed \Gianni{of 3,420 images for training, 492 
for validation, and 6,501 for testing.}%

To summarize, our contributions are as follows: \textbf{(1)} \comEmi{We introduce }MUAD: a new automotive dataset with annotations for multiple tasks and multiple uncertainty sources. \textbf{(2)} \comEmi{We perform }\comEmi{a wide} range of benchmarks on MUAD dataset \comEmi{for multiple} computer vision tasks and settings (semantic segmentation, depth estimation, object detection) to further \comEmi{support} research in this area.
\textbf{(3)} We conduct an extensive study on uncertainty quantification for 2D output tasks for recent Transformer-based architectures.

\vspace{-2mm}
\section{Related work}
\label{sec:Related}

\subsection{Datasets}
A variety of real-world datasets for autonomous driving have been recently released~\cite{cordts2016cityscapes, geiger2012we,waymo,nuscenes,yu2020bdd100k, varma2019idd,ramanishka2018toward,chang2019argoverse,houston2020one}. They have enabled tremendous progress in the area but they typically focus on a \ab{single} 
task, e.g., semantic segmentation~\cite{cordts2016cityscapes,yu2020bdd100k,ramanishka2018toward}, object detection~\cite{geiger2012we,waymo,nuscenes}, motion prediction~\cite{chang2019argoverse,houston2020one} and do not have evaluation tracks for uncertainty and out-of-distribution detection. Synthetic datasets, e.g., GTA-V \cite{richter2016playing}, SYNTHIA \cite{ros2016synthia}, virtual KITTI \cite{gaidon2016virtual} can provide abundant training data alleviating the need for costly annotation of real images \Gianni{as well as privacy preservation concerns in the case of real data.} %
Currently, they are mostly \ab{designed and} used for domain adaptation\ab{, typically imitating the content and classes from a given real dataset}. 
\ab{Several} datasets have emerged towards meeting the reliability requirement for self-driving vehicles~\cite{blum2019fishyscapes,pinggera2016lost,chan2021segmentmeifyoucan,hendrycks2019anomalyseg} and evaluate 
\ab{the performance of semantic segmentation DNNs when facing out-of-distribution objects (OOD).} Other datasets investigate the robustness against different weather conditions, e.g., night~\cite{dai2020curriculum, dai2018dark, sakaridis2021acdc}, rain~\cite{tung2017raincouver, sakaridis2021acdc}, fog~\cite{sakaridis2018semantic,sakaridis2021acdc}, however they are often acquired in different locations and conditions leading to a performance 
\ab{drop} that overlaps with the one from the difficult weather conditions. 

In order to \david{provide images of the same locations,} \ab{to} address the lack of diversity in real environments and to evaluate better the impact on the epistemic uncertainty, some works promoted 
inpainting of virtual objects~\cite{hendrycks2019anomalyseg} or synthesised weather conditions \cite{tremblay2021rain}. In this setting however, questions may be raised about the veracity of the result. \david{Therefore,} the recent ACDC dataset \cite{sakaridis2021acdc} \david{is composed entirely of real images taken from the same locations,} and includes multiple sources of aleatoric uncertainty. However, not having any control on the noise level makes it harder to quantify the link between noise and uncertainty.
Acquiring images with uncertainty corner cases is problematic as these cases are rare (long tail) and \comEmi{also} costly to annotate, e.g., 3.3 \ab{hours}/image~\cite{sakaridis2021acdc}. 
Given this scarcity, such images are better used for validation \ab{as a small test set} to assess the reliability of DNNs before deployment. 
\ab{These system validation stages can be seen as stress tests with corner cases to mirror challenging real-world conditions.}
It is thus interesting even from a more applied standpoint to have a synthetic dataset that mimics these rare conditions with some good fidelity constraint to quantify the robustness of DNNs. 
Synthetic data is abundant and can allow us to measure finer drifts in the input distribution. In addition, 
\ab{most such} datasets mainly focus on semantic segmentation, \david{while we propose to address multiple tasks (semantic segmentation, monocular depth, object detection, and instance segmentation). }

In Table \ref{table:uncertaindata} we provide a summary of the main existing uncertainty datasets. In this work, we propose a fully synthetic dataset, called MUAD, integrating different weather conditions with various intensities, and suitable for a multitude of vision tasks and for the comprehensive characterisation of their uncertainty.

\begin{table}[t]
\setlength{\abovecaptionskip}{0.cm}
\begin{center}
\scalebox{0.52}
{
\begin{tabular}{l|rccccrcccc} 
\toprule
\textbf{Dataset} & \multicolumn{1}{l}{\begin{sideways}\begin{tabular}[c]{@{}l@{}}\textbf{Adversarial}\\\textbf{annotations}\end{tabular}\end{sideways}} & \begin{sideways}\textbf{Fog}\end{sideways} & \begin{sideways}\textbf{Night}\end{sideways} & \begin{sideways}\textbf{Rain}\end{sideways} & \begin{sideways}\textbf{Snow}\end{sideways} & \multicolumn{1}{r}{\begin{sideways}\textbf{Classes}\end{sideways}} & \begin{sideways}\begin{tabular}[c]{@{}l@{}}\textbf{Out of}\\\textbf{distribution}\end{tabular}\end{sideways} & \begin{sideways}\textbf{Depth}\end{sideways} & \begin{sideways}\begin{tabular}[c]{@{}l@{}}\textbf{Object}\\\textbf{detection}\\\textbf{2D/3D}\end{tabular}\end{sideways} & \begin{sideways}\begin{tabular}[c]{@{}l@{}}\textbf{Instance}\\\textbf{segmentation}\end{tabular}\end{sideways} \\  \midrule
Foggy Driving\cite{sakaridis2018semantic}        & 101                                                   & \checkmark & -                         & -                         & -                         & 19                                    & -                            & -                         & \Xuanlong{\checkmark}                         & -                              \\ \midrule
Foggy Zurich \cite{dai2020curriculum}            & 40                                                    & \checkmark & -                         & -                         & -                         & 19                                    & -                            & -                         & -                         & -                              \\ \midrule
Nighttime Driving \cite{dai2018dark}             & \Xuanlong{50}                                                   & -                         & \checkmark & -                         & -                         & 19                                    & -                            & -                         & -                         & -                              \\ \midrule
Dark Zurich \cite{sakaridis2020map}              & 201                                                   & -                         & \checkmark & -                         & -                         & 19                                    & -                            & -                         & -                         & -                              \\ \midrule
Raincouver \cite{tung2017raincouver}             & 326                                                   & -                         & \checkmark & \checkmark & -                         & 3                                     & -                            & -                         & -                         & -                              \\ \midrule
WildDash \cite{zendel2018wilddash}               & 226                                                   & \checkmark & \checkmark & \checkmark & \checkmark & 19                                    & -                            & -                         & -                         & -                              \\ \midrule
BDD100K \cite{yu2020bdd100k}                     & 1346                                                  & \checkmark & \checkmark & \checkmark & \checkmark & 19                                    & -                            & -                         & -                         & -                              \\ \midrule
ACDC \cite{sakaridis2021acdc}                    & 4006                                                  & \checkmark & \checkmark & \checkmark & \checkmark & 19                                    & -                            & \checkmark & \checkmark & -                              \\ \midrule
\Xuanlong{Virtual KITTI 2~\cite{cabon2020vkitti2}}& 21260 & \checkmark & - & \checkmark & -  & 14 & - & \checkmark & \checkmark & \checkmark\\ \midrule
Fishyscapes \cite{blum2019fishyscapes}           & \multicolumn{1}{r}{373}                                 & -                         & -                         & -                         & -                         & \multicolumn{1}{r}{19+2}             & \checkmark    & -                         & -                         & -                              \\ \midrule
LostAndFound \cite{pinggera2016lost}             & \multicolumn{1}{r}{1203}                                 & -                         & -                         & -                         & -                         & \multicolumn{1}{r}{19+9}             & \checkmark    & -                         & -                         & -                              \\ \midrule
RoadObstacle21  \cite{chan2021segmentmeifyoucan} & 327                                                   & -                         & \checkmark                         & -                         & \checkmark                         & \multicolumn{1}{r}{19+1}                 & \checkmark    & -                       & -                         & -                              \\ \midrule
RoadAnomaly21 \cite{chan2021segmentmeifyoucan}   & 100                                                   & -                         & -                         & -                         & \checkmark                         & \multicolumn{1}{r}{19+1}                 & \checkmark    & -                       & -                         & -                              \\ \midrule
Streethazard~\cite{hendrycks2019anomalyseg}                                                      & 6625                                                  & -                         & -                         & -                         & -                         & \multicolumn{1}{r}{13+250}                 & \checkmark    & -                         & -                         & -                              \\ \midrule
BDD anomaly~\cite{hendrycks2019anomalyseg}                                                       & 810                                                   & \checkmark                          & \checkmark                       & \checkmark                         & \checkmark                          & \multicolumn{1}{r}{17+2}             & \checkmark    & -                         & -                         & -                              \\ \midrule
MUAD & 10413 & \checkmark & \checkmark & \checkmark & \checkmark  & 16+9 & \checkmark    & \checkmark & \checkmark & \checkmark\\

\bottomrule

\end{tabular}
} %
\end{center}
\vspace{-3mm}
\caption{\small \textbf{Comparative overview} of the different datasets for uncertainty on autonomous driving.  }
\label{table:uncertaindata}
\vspace{-2mm}
\end{table}

\subsection{Uncertainty}

Several works address the two types of uncertainty, in particular for the classification task.
Most approaches build upon Bayesian learning, frequently \david{using} Bayesian Neural Networks (BNNs)~\cite{blundell2015weight,hernandez2015probabilistic,maddox2019simple,franchi2020tradi,wilson2020bayesian,dusenberry2020efficient,franchi2020encoding}, which estimate the posterior distribution of the DNN weights to marginalize the likelihood distribution at inference time. Yet \ab{most} BNNs are difficult to train and scale to complex computer vision tasks \ab{ that have been addressed, so far, by fewer uncertainty estimation methods}. 
\ab{Ensembles~\cite{lakshminarayanan2017simple} and pseudo-ensembles~\cite{gal2016dropout,maddox2019simple,franchi2020tradi} achieve state-of-the-art performance on various tasks, at the high cost of multiple training and/or multiple forward passes at inference.}
Some approaches \cite{kendall2017uncertainties} formalize DNNs to output a parametric distribution, and their goal is to estimate the distribution parameters. These approaches can be applied to optical flow~\cite{ilg2018uncertainty} and object detection~\cite{choi2019gaussian}, yet they mainly focus on aleatoric uncertainty.
\ab{Besides the additional challenges posed by complex computer vision tasks, progress on uncertainty estimation in this area is hindered by the lower number of datasets for properly assessing both the quality of the predictive uncertainty and the predictive performance. With MUAD we hope to encourage research in this essential area for practical applications with annotations and benchmarks for multiple tasks.}

\section{Multiple Uncertainties for Autonomous Driving benchmark (MUAD)}
\label{sec:Datasets}

\begin{figure}[t!]
\setlength{\abovecaptionskip}{0.cm}
    \centering{\includegraphics[width=0.6\linewidth]{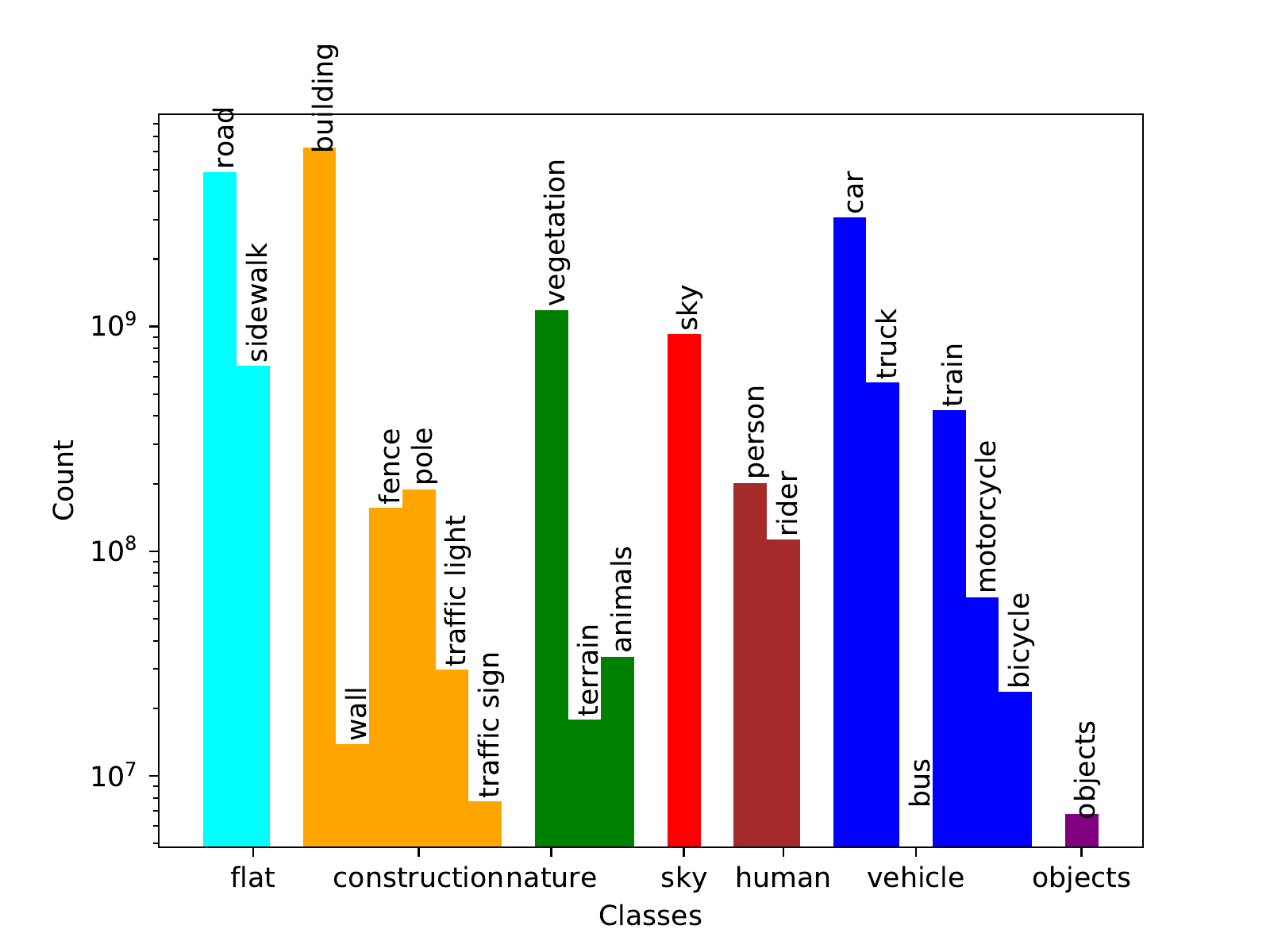}}
        \vspace{-1.5em}
    \caption{\small \textbf{Number of  annotated pixels per class in MUAD.}} %
    \label{fig:stats}
    \vspace{-1em}
\end{figure}

\Xuanlong{According to the 
categorization of the uncertainty in line with 
\comEmi{the current works of the community~\cite{gawlikowski2021survey} (summarized in the Supplementary Material),}
we propose to use the dataset to better evaluate the results and uncertainty estimations given by the DNNs in the context of autonomous driving. \comEmi{Let us link the two main types of uncertainty - aleatoric and epistemic - to the specific context of our application.} In the scenario of autonomous driving, we believe that the aleatoric uncertainty of the DNNs will \comEmi{occur} due to different weather conditions than the ones \comEmi{present} in the training set. The epistemic uncertainty of the DNNs should arise when the class or the \comEmi{appearance} of objects in the picture \comEmi{differ from those of} the data provided in the training set. The design of MUAD dataset is based on 
\david{
this hypothesized} relationship between uncertainty and autonomous driving scenarios. In the remainder of this section, we will detail the composition of MUAD dataset.}

\Emi{
The goal of MUAD is to confront DNNs to uncertain environments and to characterize numerically their robustness in adverse conditions, more specifically in the presence of rain, fog and %
snow. Photorealism is essential for guaranteeing that synthetic datasets are challenging with respect to real-world conditions, and also for keeping them relevant for use in industrial applications. This is particularly important for accommodating weather artifacts~\cite{li2017photo,von2019simulating,tremblay2021rain}. \AnyV{Our dataset is generated using a physics based synthetic image rendering engine to produce 
\ab{high-quality realistic images and sequences.} The engine uses an accurate light transport model~\cite{veach1998robust, magnor2015digital} and provides a physics
description of lights, cameras and materials. This allows for a 
detailed simulation of the
amount of light that is reaching the camera sensor. The camera sensor itself is simulated converting the energy coming from the scene in the form of photons into electrons. Electrons are finally converted into a voltage that is digitized to produce the digital values that represent the color image.} \Xuanlong{We provide the photorealistic rendering descriptions for different weather conditions in Section \ref{sec:rendering}.} For each sample \Xuanlong{in MUAD}, the corresponding ground truth information contains the semantic segmentation, the depth map, and for some specific classes 
(pedestrian, car, van, traffic light, traffic sign) 
the instance segmentation with the corresponding bounding boxes. We follow the standard data split strategy, however the training and validation set contain only images with normal weather conditions and without some specific classes which are denoted as OOD. The test set is organized into seven subsets following the intensity of the \comEmi{adverse} weather conditions: 
\begin{itemize}[noitemsep,topsep=0pt,parsep=0pt,partopsep=0pt]
\item \textbf{normal set:} images without OOD objects nor \comEmi{adverse} conditions, as in Figure \ref{fig:teaser}a.
\item \comEmi{\textbf{normal set overhead sun:} images without OOD objects nor adverse conditions, in which we simulate the sun with a zenith angle of 0$^{\circ}$, that we denote for the sake of simplicity as overhead sun.}
\item \textbf{OOD set:} images 
\ab{with} OOD objects and without \comEmi{adverse} conditions, as in Figure \ref{fig:teaser}b.
\item \textbf{low adv. set:} images 
\ab{with} medium intensity \comEmi{adverse} conditions (%
fog, rain or snow).%
\item \textbf{high adv. set:} images containing high intensity \comEmi{adverse} conditions (%
fog, rain or snow).%
\item \textbf{low adv. with OOD set:} images containing both OOD objects and medium intensity \comEmi{adverse} conditions (%
fog, rain or snow), as in Figure \ref{fig:teaser}c. 
\item \textbf{high adv. with OOD  set:} images containing both OOD objects and high intensity \comEmi{adverse} conditions (%
fog, rain or snow), as in Figure \ref{fig:teaser}d. 
\end{itemize}
}

\begin{figure*}[!t]
     \centering
     \subfigure[]{\includegraphics[width=0.3\linewidth]{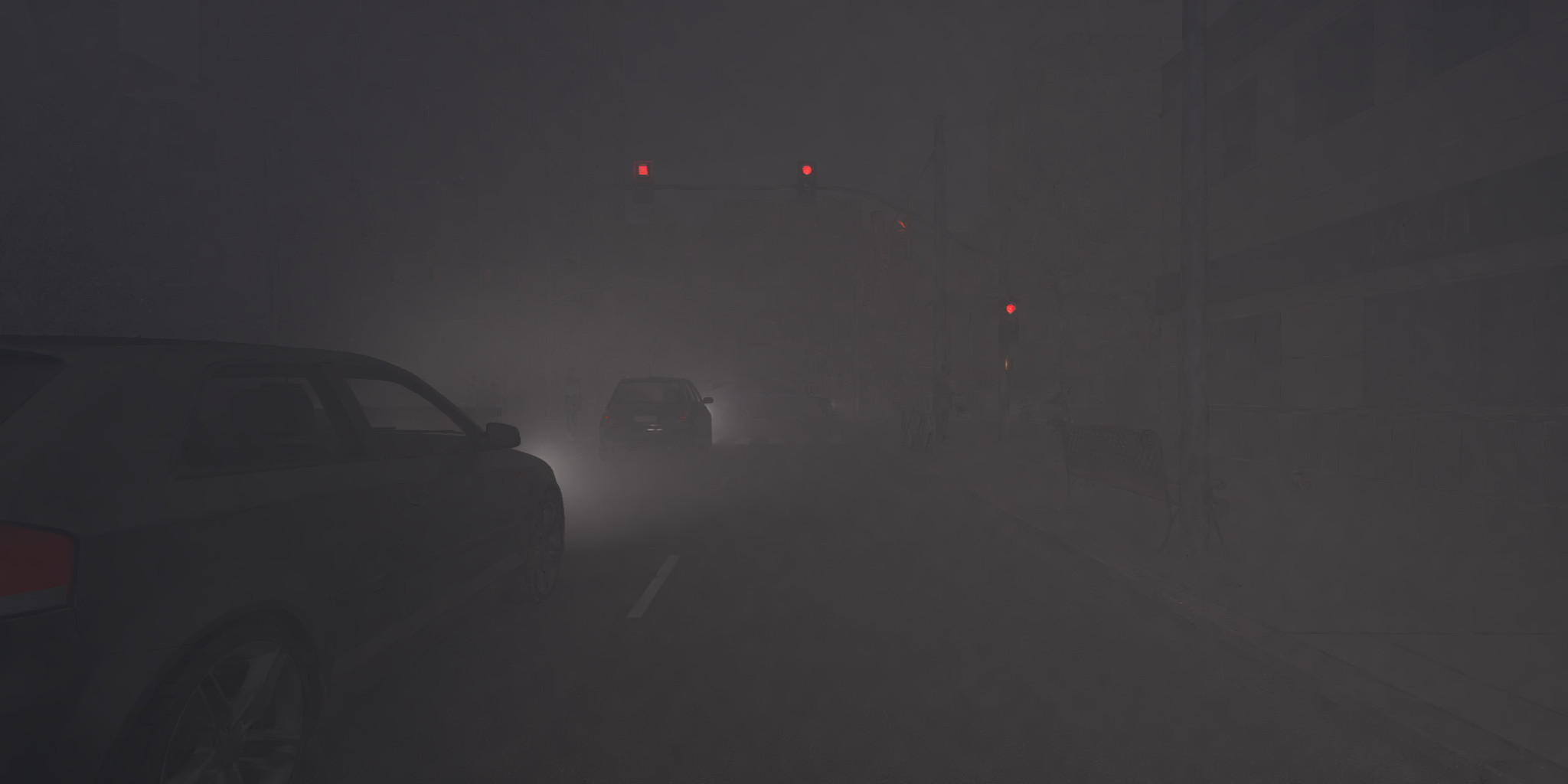}\label{fog_img}}
    \subfigure[]{\includegraphics[width=0.3\linewidth]{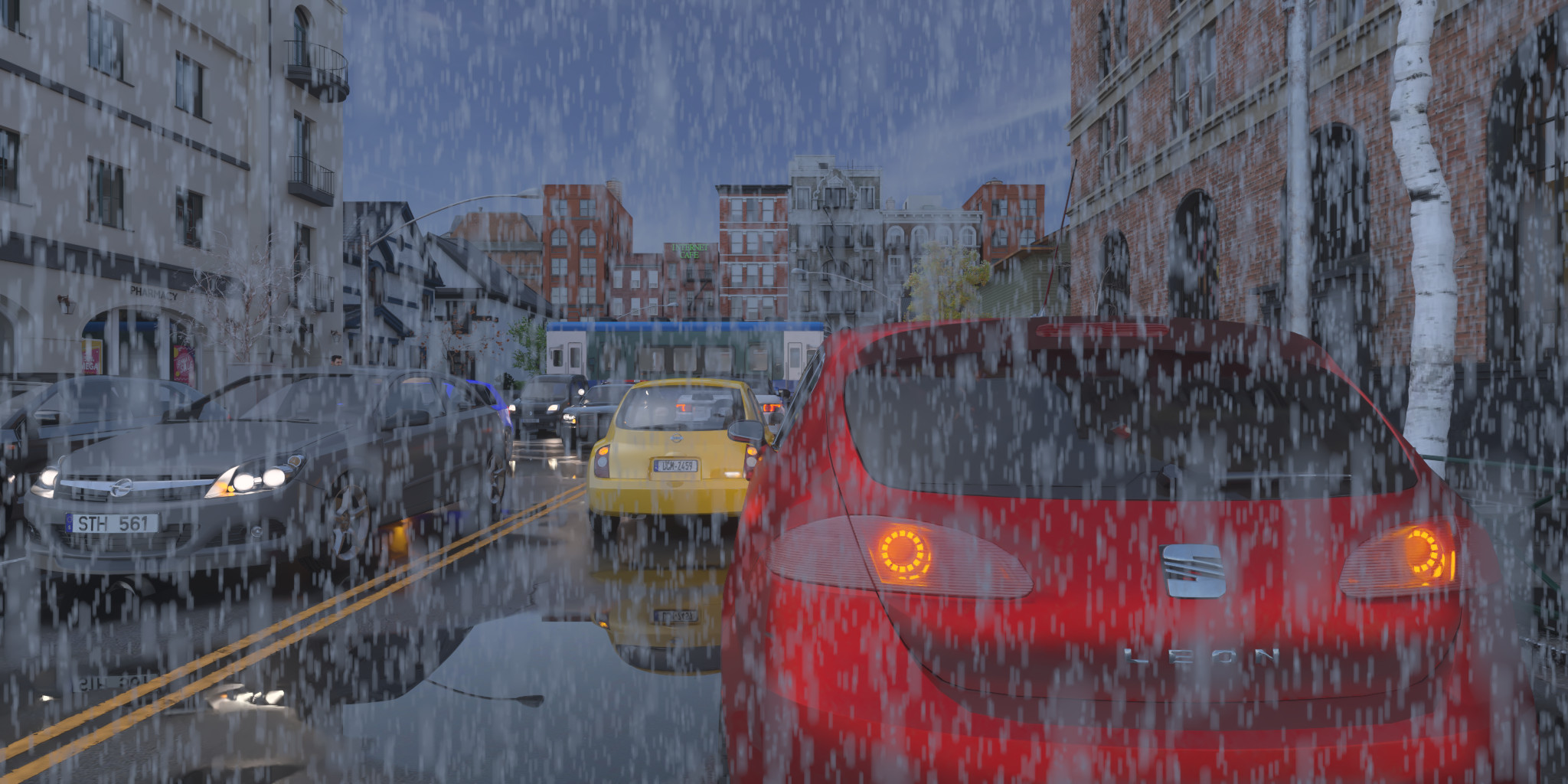}\label{rain_img}}
    \subfigure[]{\includegraphics[width=0.3\linewidth]{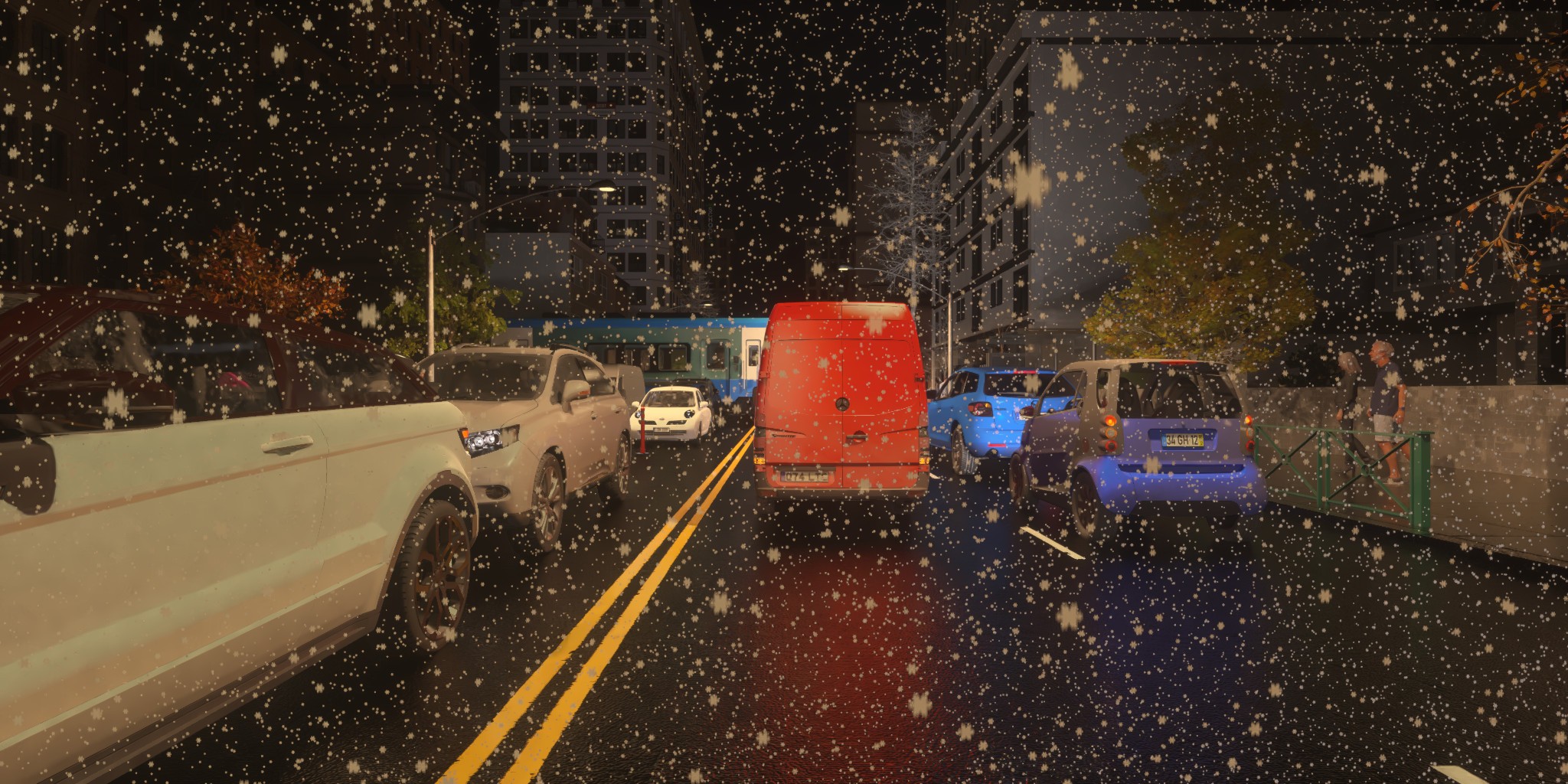}\label{snow_img}}\\
    \subfigure[]{\includegraphics[width=0.3\linewidth]{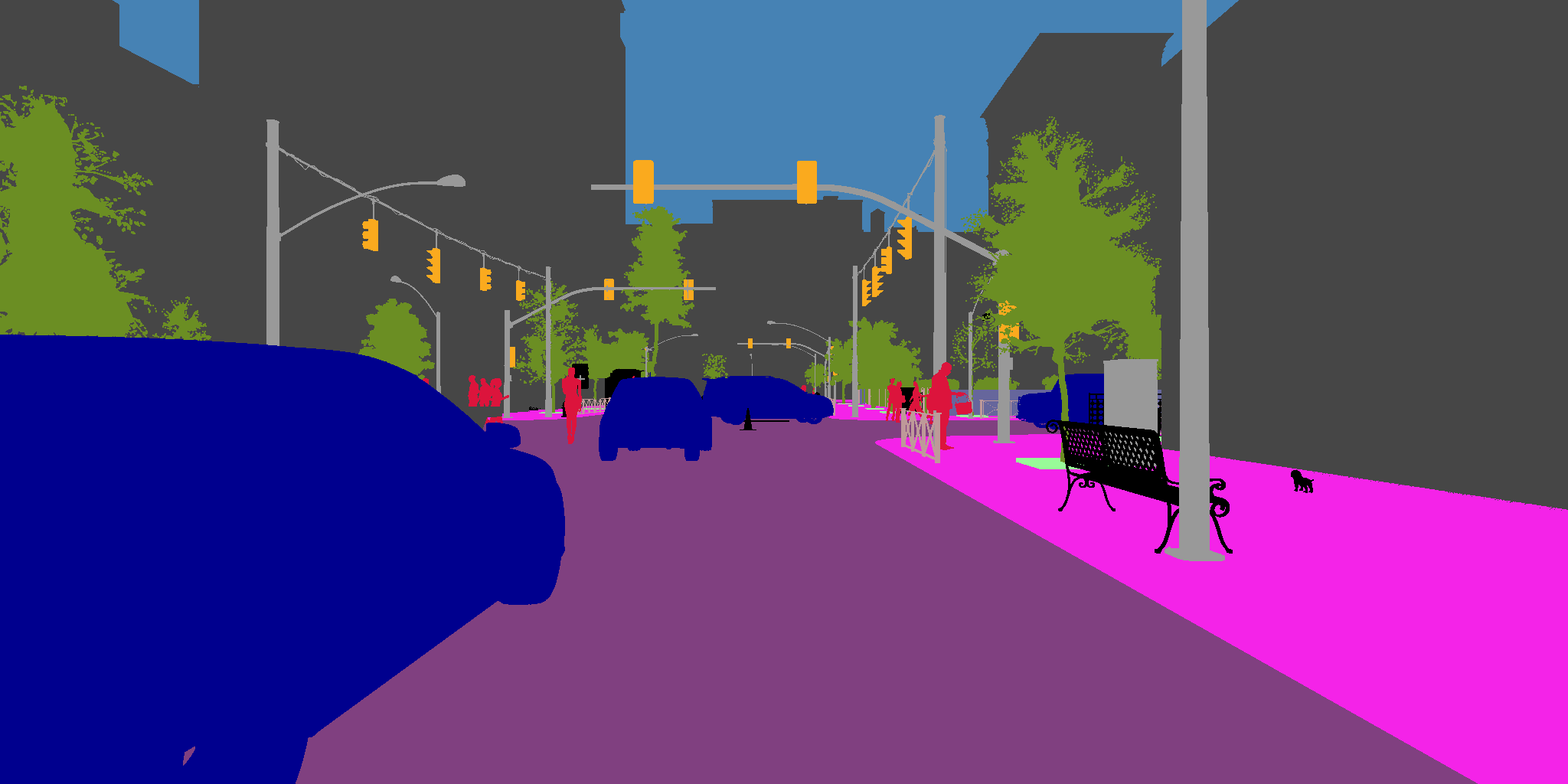}\label{fog_seg}}
    \subfigure[]{\includegraphics[width=0.3\linewidth]{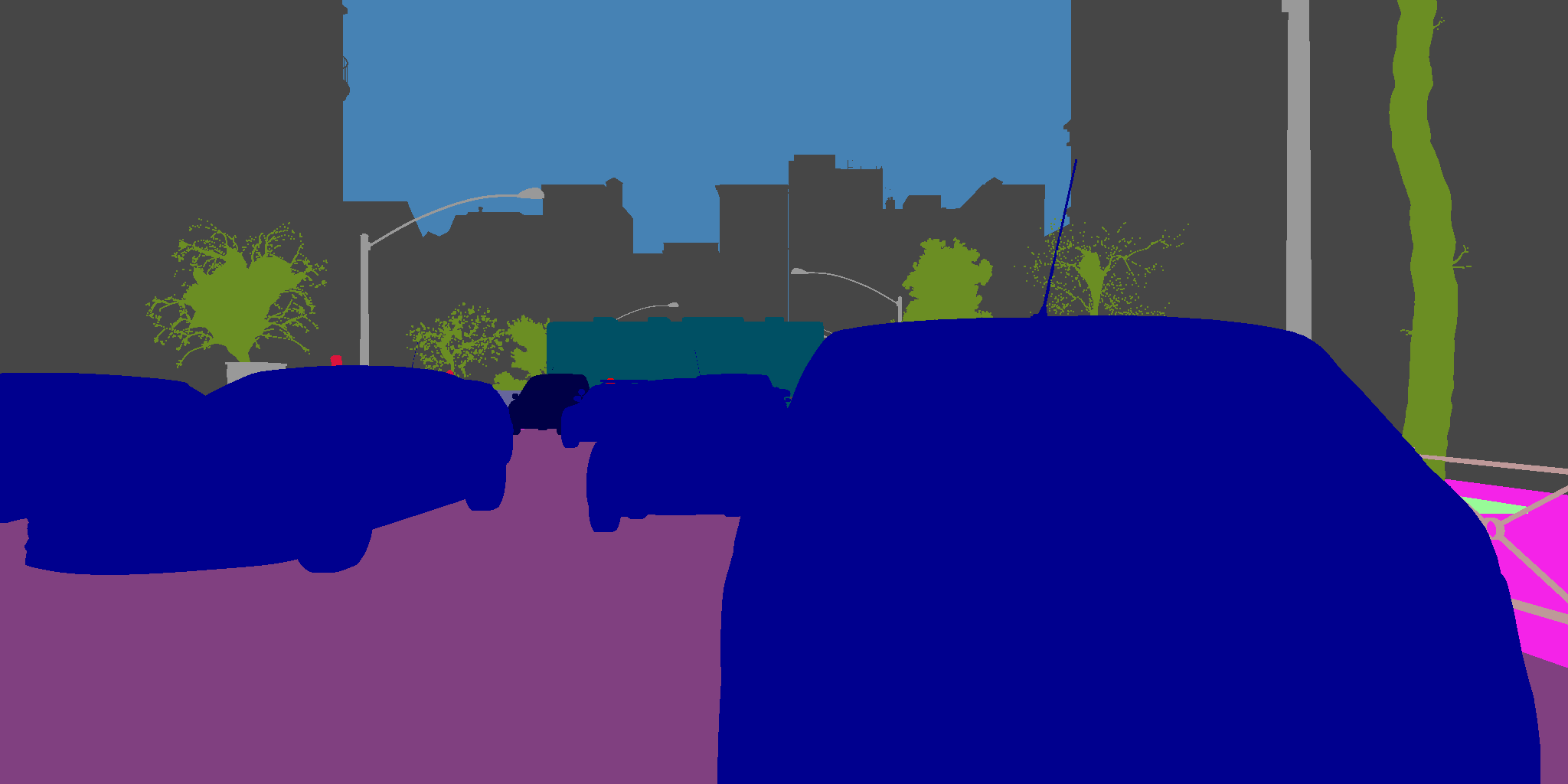}\label{rain_seg}}
    \subfigure[]{\includegraphics[width=0.3\linewidth]{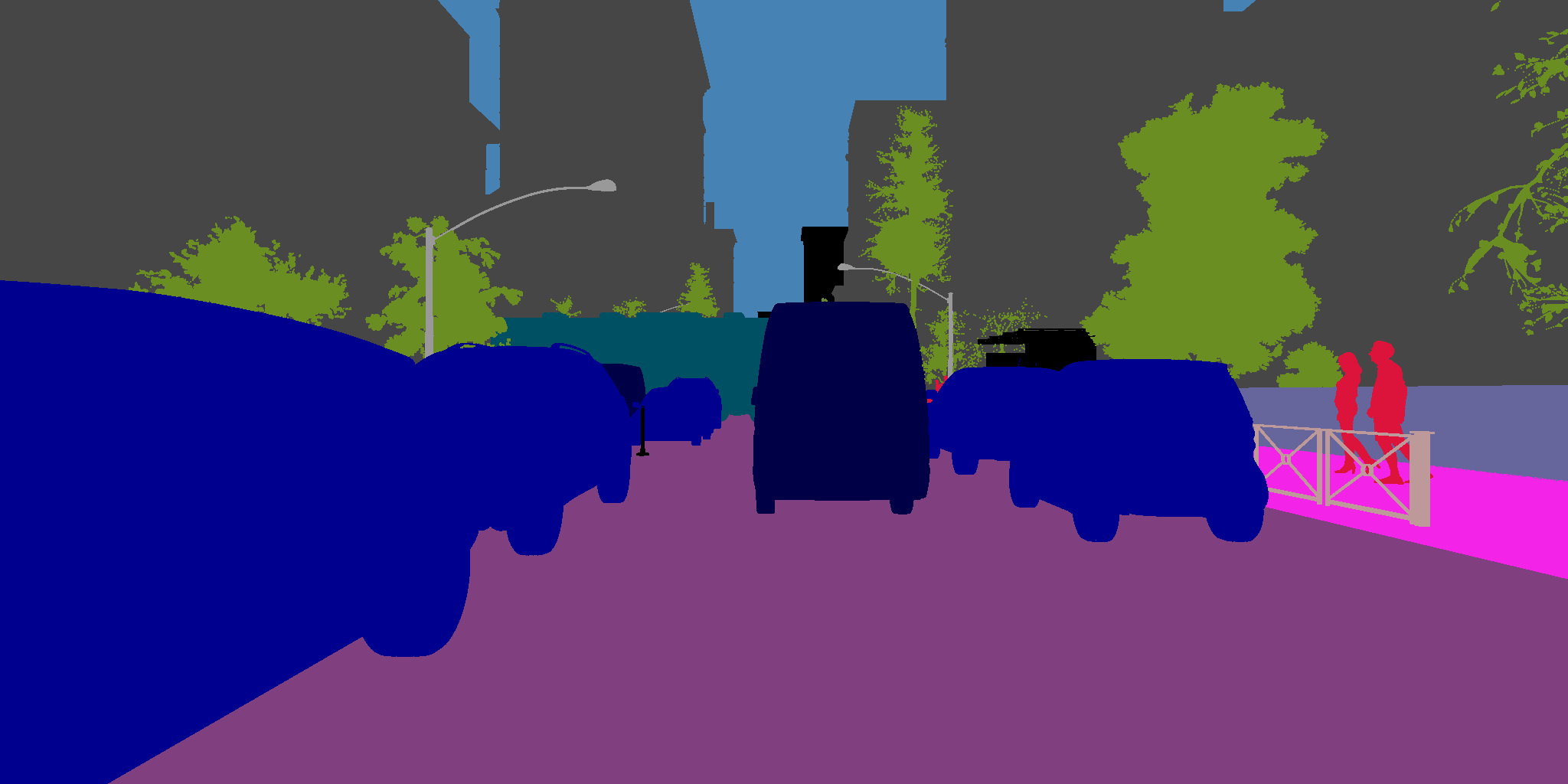}\label{snow_seg}}
    \caption{\small \textbf{Illustration of semantic segmentation images of MUAD dataset.} The first row is composed of the original images of the \textbf{high adv. set}. The second row is their corresponding ground truth.}
    \vspace{-3mm}
     \label{fig:data_seg}
\end{figure*}

\begin{figure*}[!t]
     \centering
     \subfigure[]{\includegraphics[width=0.3\linewidth]{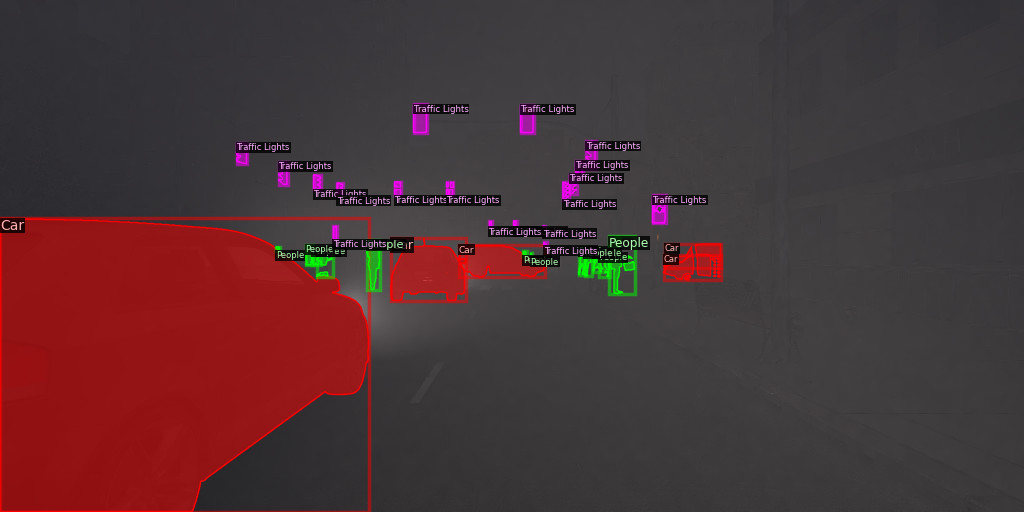}\label{fog}}
    \subfigure[]{\includegraphics[width=0.3\linewidth]{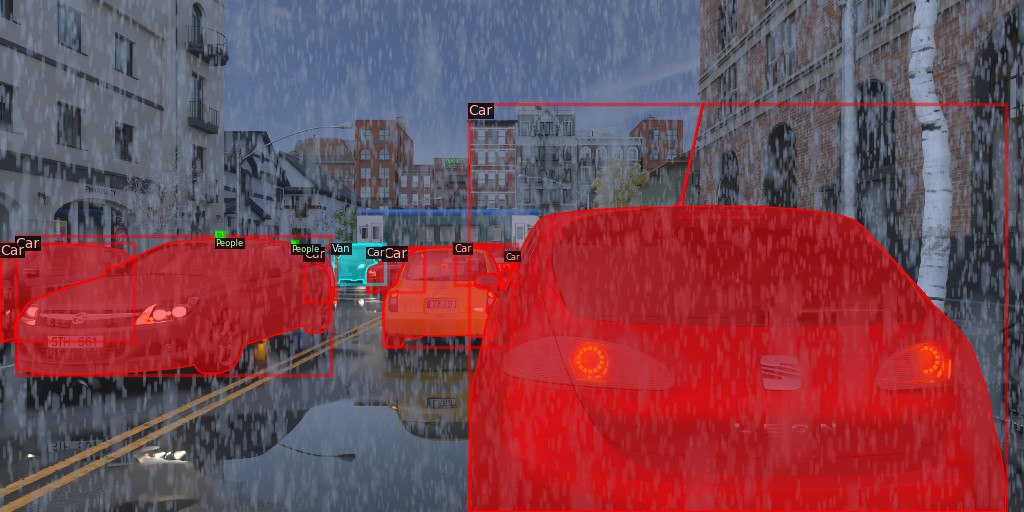}\label{rain}}
    \subfigure[]{\includegraphics[width=0.3\linewidth]{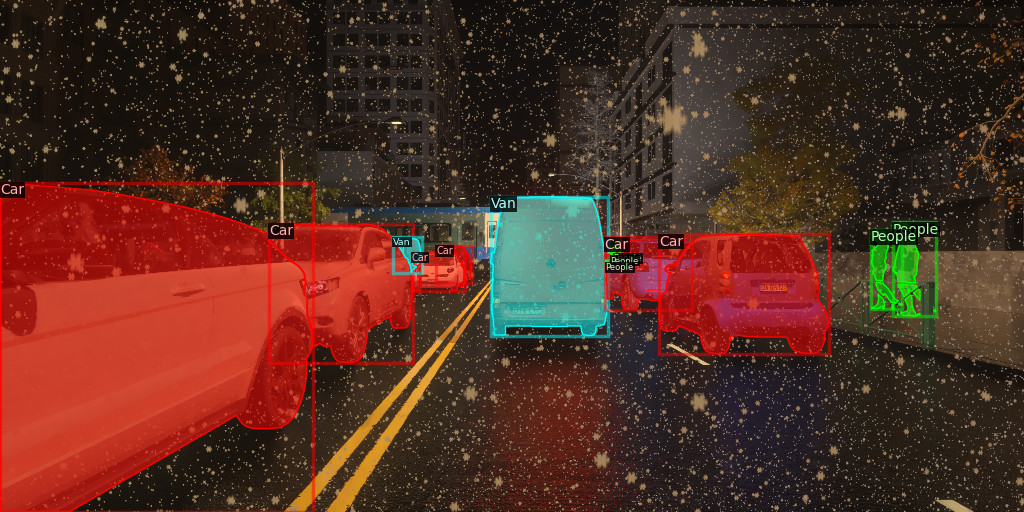}\label{snow}}
    \caption{\small \textbf{Illustration of instance segmentation images of MUAD dataset.}
The three images are selected from the \textbf{high adv. set}. We illustrated fog (\ref{fog}), rain  (\ref{rain}), and snow (\ref{snow}) conditions.}
     \label{fig:data}
\end{figure*}

In Figure \ref{fig:data_seg} and \ref{fig:data} we illustrate the instance segmentation and the semantic segmentation of 3 images. 
\ab{
The adverse weather conditions are realistic and challenging as they bring a mix of difficult (unknown during training) environment conditions and perturbation of the visibility in the scene}. 
\ab{We argue that such settings are helpful for autonomous driving since the autonomous system must face and be robust against a variety of weather conditions and situations.}

\begin{table}[t!]
\setlength{\abovecaptionskip}{0.cm}
\centering
\scalebox{0.5}{\setlength{\tabcolsep}{1mm}{
\begin{tabular}{llr} 
\toprule
\begin{tabular}[c]{@{}l@{}}\textbf{Cityscapes}\\\textbf{classes}\end{tabular} & \multicolumn{1}{c}{\textbf{MUAD classes}} & \multicolumn{1}{l}{\begin{tabular}[c]{@{}l@{}}\textbf{nb. of images with}\\\textbf{the annotations}\end{tabular}} \\ 
\midrule
Road & \begin{tabular}[c]{@{}l@{}}Bots, Tram Tracks, Crosswalk, Parking Area, Garbage - Road, \\Road Lines, Sewer Longitudinal Crack, Transversal Crack, Road, Asphalt hole,\\Polished Aggregate, Vegetation - Road, Sewer - Road, Construction Concrete\end{tabular} & 9,055 \\ 
\midrule
Sidewalk & Lane Bike, Kerb Stone, Sidewalk, Kerb Rising Edge & 8,948 \\ 
\midrule
Building & \begin{tabular}[c]{@{}l@{}}House, Construction Scaffold, Building, Air Conditioning, Construction Container,\\ TV Antenna, Terrace, Water Tank, Pergola Garden, Stairs, Dog House,\\ Sunshades, Railings, Construction Stock, Marquees, Hangar Airport\end{tabular} & 9,089 \\ 
\midrule
Wall & Wall & 1,101 \\ 
\midrule
Fence & Construction Fence, Fences & 8,622 \\ 
\midrule
Pole & Traffic Signs Poles or Structure, Traffic Lights Poles, Street lights, Lamp & 8,984 \\ 
\midrule
Traffic light & Traffic Lights Head, Traffic Cameras, Traffic Lights Bulb (red, yellow, green) & 8,222 \\ 
\midrule
Traffic sign & Traffic Signs & 2,672 \\ 
\midrule
Vegetation & Vegetation & 9,072 \\ 
\midrule
Terrain & Terrain, Tree Pit & 8,377 \\ 
\midrule
Sky & Sky & 8,591 \\ 
\midrule
Person & Walker, All colors of Construction Helmet, All colors of Safety Vest, Umbrella, People & 8,843 \\ 
\midrule
Rider & Cyclist, Biker & 3,470 \\ 
\midrule
Car & Car, Beacon Light, Van, Ego Car & 9,026 \\ 
\midrule
Truck & Truck & 5,533 \\ 
\midrule
Bus & Bus & 0 \\ 
\midrule
Train & Train, Subway & 2,240 \\ 
\midrule
Moto\comEmi{r}cycle & Motorcycle, Segway, Scooter Child & 2,615 \\ 
\midrule
Bicycle & Bicycle, Kickbike, Tricycle & 2,816 \\
\midrule
Animals & Cow, Bear, Deer, Moose & 603 \\ 
\midrule
Objects anomalies & 
\comEmi{Food Stand}, 
Trash Can, Garbage Bag & 352 \\ 
\midrule
Background & Others & - \\
\bottomrule
\\
\end{tabular}
}}
\vspace{-1mm}
\caption{\small{Overview of annotated classes}}
\vspace{-3mm}
\label{Overview}
\end{table}

\subsection{MUAD statistics}
Our dataset %
contains 3,420 images in the train set, and 492 in the validation set. The test set is composed of 6,501 %
images divided as follows: 551 in the \textbf{normal set}, \Xuanlong{102} in the \textbf{normal set no shadow,} \Xuanlong{1,668} in the \textbf{OOD set}, \Xuanlong{605} in the \textbf{low adv. set},  \Xuanlong{602} in the \textbf{high adv. set}, 1,552 in the \textbf{low adv. with OOD set} and 1,421 in the \textbf{high adv. with OOD set}. {All of these sets cover day and night conditions with 2/3 of day images and 1/3 of night images. }%
Test datasets address diverse weather conditions (rain, snow, and fog with different levels), \Xuanlong{ and various OOD objects. The resolution of all images is 1024\ab{$\times$}2048.}  %

 The dataset aims to provide a general and consistent coverage for a typical urban and suburban environment under different times of day and weather conditions. \AnyV{Ego-vehicle poses are drawn randomly within a complex environment, and in a second stage the field of view is populated stochastically with dynamic objects of interest following distributions in compliance with their expected behaviour}. The pose and context changes as well as the variation of the models for the objects of interest ensure that content diversity is high, in addition to images being photorealistic. \Gianni{The simulator makes use of approximately 300 different person models and 150 different vehicle models, which are sampled while varying their visual characteristics.}

\subsection{Class labels}

\Gianni{The class ontology of MUAD is presented in Table \ref{Overview}. MUAD comprises 155 different classes that we have regrouped into 21 classes. The first 19 classes are similar to the CityScapes classes \cite{cordts2016cityscapes}, then we added object anomalies and animals to have more diversity in the anomalies. }
\Emi{In addition to ensuring high content diversity, this ontology 
facilitates the mapping of MUAD to specific environments which require or impose a lower number of more generic classes. Consequently, trained models are easily transferable for existing datasets, and we provide the mapping towards the 21 classes widely used by the community, e.g., \cite{chen2017no, cordts2016cityscapes,ros2016synthia,richter2016playing}.  The dataset statistics for the 21 %
classes are presented in Figure \ref{fig:stats}.
}\Gianni{For the evaluation of OOD detection, we have excluded nine classes (train, motorcycle, bicycle, bears, cow, deers, moose, 
\comEmi{food stand},
garbage bags) from the training and validation sets. These classes are present in the test set as OOD objects. DNNs that 
\comEmi{process samples belonging to }one of these nine %
classes are expected to have a low confidence score.}

\subsection{Photorealistic rendering}
\label{sec:rendering}

\Xuanlong{Our physically based approach simulates the weather conditions taking into consideration the amount of ozone, the humidity, among
other factors.} \Gianni{\textbf{Regarding the sky}, the renderer uses a physical model of the light coming from the sky. The amount of ozone and humidity in the atmosphere changes the emissive spectral profile of the sky, impacting the color of the objects in the scene. 
Apart from ozone and humidity, there are other factors that the render takes into account, for instance, turbidity and scattering asymmetry. \textbf{Regarding the rain and the snow},
the simulation of every raindrop allows us to model physical dispersion. 
For improved realism
we choose the falling speed and size of raindrops according to observed real rain \cite{muramoto1989measurement,baranidesign1}. For snow, the same principle applies, but changing in this case the
material and the dynamics. \textbf{Regarding the fog}, we use a full volumetric approach for the simulation where scattering effects are considered. \textbf{Regarding the level of noise}, to the best of our knowledge, there is no standard procedure to measure the intensity of adverse weather conditions for driving scenarios. We empirically selected the number of raindrops, snowflakes, and fog intensity from a human point or view. All the efforts mentioned above improved our dataset realism. A study \cite{Anyverse1} was performed that confirmed that our render enhances the realism of MUAD compared to SYNTHIA.}

\section{Experiments}

\subsection{Semantic segmentation experiments}

\begin{table*}[t]
\setlength{\abovecaptionskip}{0.cm}
\begin{center}
 \scalebox{0.40}
 {
\begin{tabular}{llcc|cc|cc} 
\toprule
\multicolumn{1}{l}{\multirow{2}{*}{Methods}} & \multicolumn{1}{l}{\multirow{2}{*}{Architectures}}& \multicolumn{2}{c}{\textbf{normal set}}  & \multicolumn{2}{c}{\textbf{low adv. without OOD set}}  & \multicolumn{2}{c}{\textbf{high adv. without OOD set}}  \\
   ~ &   ~ & mIoU $\uparrow$ &  ECE $\downarrow$   &  mIoU $\uparrow$  &  ECE $\downarrow$   &  mIoU $\uparrow$ &  ECE $\downarrow$   \\
\midrule
Baseline (MCP)~\cite{hendrycks2016baseline} &  DeepLab v3+ ~\cite{chen2018encoder} &  68.90\% & 0.0138 & 38.77\% & 0.3238 & 22.51\% & 0.4567\\
\midrule
Baseline (MCP)~\cite{hendrycks2016baseline} &  SegFormer-B0  \cite{xie2021segformer} &  69.04\% & 0.0135 & \second 48.67\% & 0.1004 & \second 30.0\% & \second 0.2396\\
\midrule
MC-Dropout \cite{gal2016dropout} &  DeepLab v3+ ~\cite{chen2018encoder} &  65.33\% & 0.0172 & 42.08\% & 0.2587 & 27.68\% & 0.3846\\
 \midrule
MC-Dropout \cite{gal2016dropout} &  SegFormer-B0  \cite{xie2021segformer} & 68.55\% & \second 0.0119 & 45.01\% & \first \textbf{0.0758} & 26.59\% & \first \textbf{0.1594} \\
\midrule
Deep Ensembles ~\cite{lakshminarayanan2017simple} &  DeepLab v3+ ~\cite{chen2018encoder} & \second 69.80\% & 0.0129 & 42.81\% & 0.2444 & 23.91\% & 0.4500\\
 \midrule
Deep Ensembles ~\cite{lakshminarayanan2017simple} &  SegFormer-B0  \cite{xie2021segformer} &  \first \textbf{70.00\%} & \first \textbf{0.0115} & \first \textbf{49.10\%} & \second 0.0837 & \first \textbf{31.67\%} & 0.3167 \\

\bottomrule
\end{tabular}
}

 \scalebox{0.4}
 {
\begin{tabular}{ll|ccccc|ccccc|ccccc} 
\toprule
\multicolumn{1}{l}{\multirow{2}{*}{Methods}} & \multicolumn{1}{l}{\multirow{2}{*}{Architectures}} & \multicolumn{5}{c}{\textbf{OOD set}}  & \multicolumn{5}{c}{\textbf{low adv. with OOD set}}  & \multicolumn{5}{c}{\textbf{high adv. with OOD set}}  \\
    & ~  &  mIoU $\uparrow$  &  ECE $\downarrow$ &  AUROC $\uparrow$ &  AUPR $\uparrow$ &  FPR $\downarrow$  &  mIoU $\uparrow$ &  ECE $\downarrow$  &  AUROC $\uparrow$ &  AUPR $\uparrow$ &  FPR $\downarrow$   & mIoU $\uparrow$    & ECE $\downarrow$ &   AUROC $\uparrow$ &  AUPR $\uparrow$ &  FPR $\downarrow$  \\
\midrule
Baseline (MCP)~\cite{hendrycks2016baseline} &    DeepLab v3+ ~\cite{chen2018encoder} & 57.32\% & 0.0607 & 0.8624 & 0.2604 & 0.3943 & 31.84\% & 0.3078 & 0.6349 & 0.1185 & 0.6746 & 18.94\% & 0.4356 & 0.6023 & 0.1073 & 0.7547\\
\midrule
Baseline (MCP)~\cite{hendrycks2016baseline} &   SegFormer-B0  \cite{xie2021segformer} & \second 58.91 \% &  0.06465 & 0.8578 & 0.21576 & 0.4106 & \first \textbf{40.22\%} & 0.1544 & 0.7448 & 0.1361 & 0.5876 & \first \textbf{27.51 \%} & 0.6564 &  0.6564 & 0.1071 & 0.7011\\
\midrule
MC-Dropout \cite{gal2016dropout}&  DeepLab v3+ ~\cite{chen2018encoder} & 55.62\% & 0.0645 & 0.8439 & 0.2225 & 0.4575 & 33.38\% & \second 0.1329 &  \second 0.7506 &   \second 0.1545 &  \second 0.5807 &   20.77\% &  0.3809 &  \second 0.6864 &  \second 0.1185 &  \second 0.6751\\
 \midrule
MC-Dropout \cite{gal2016dropout}&  SegFormer-B0  \cite{xie2021segformer} & 58.81\% & \first \textbf{0.0574} & \second 0.8811 & 0.2535 & \second 0.3435 & 39.64\% & \first \textbf{0.1172} & \first \textbf{0.7698} & \first \textbf{0.1557} & \first \textbf{0.5498} & \second 26.52\% & \first \textbf{0.1771} & \first \textbf{0.6965} & \first \textbf{0.1237} & \first \textbf{0.6633}\\
 \midrule
Deep Ensembles ~\cite{lakshminarayanan2017simple} &  DeepLab v3+ ~\cite{chen2018encoder}   & 58.29\% &  \second 0.0588 &   0.871 &  \first \textbf{0.2802} &  0.3760 &  34.91\% &  0.2447 & 0.6543 & 0.1212 & 0.6425 & 20.19\% &  0.4227 & 0.6101 & 0.1162 & 0.7212\\
 \midrule
 Deep Ensembles ~\cite{lakshminarayanan2017simple} &  SegFormer-B0  \cite{xie2021segformer}   & \first \textbf{59.50\%} & 0.05928 & \first \textbf{0.8843} & \second 0.2611 & \first \textbf{0.3342} & \second 40.00 \% & 0.1400 & 0.6933 & 0.1198 & 0.6290 & 25.89 \%& \second 0.3305 & 0.5939 & 0.0959 & 0.7287\\
\bottomrule
\end{tabular}
}
\end{center}
\vspace{-2mm}
\caption{\small\textbf{{Comparative results for semantic segmentation on MUAD}}. {The mIoU is related to the main task performance, while the rest of the metrics evaluate the uncertainty quality when the model is confronted with different types of perturbations.}
}
\label{table:tnb2}
\vspace{-2mm}
\end{table*}

Our semantic segmentation study consists of two experiments.  Firstly, we evaluate on MUAD the uncertainty quantification of three 
benchmarks
(MCP~\cite{hendrycks2016baseline}, Deep Ensembles ~\cite{lakshminarayanan2017simple}, MC Dropout \cite{gal2016dropout}), by taking advantage of the OOD/adverse weather splits. 
The second experiment evaluates the quality of transfer learning from MUAD to Cityscapes~\cite{cordts2016cityscapes} and the quality of the uncertainty quantification on Cityscapes. We aim here to verify whether MUAD can be used for unsupervised domain adaptation.

\david{For the first experiment, }\ab{we train a DeepLabV3+~\cite{chen2018encoder} network with ResNet50 encoder \cite{he2016deep} and a SegFormer-B0 \cite{xie2021segformer} on MUAD.} %
Table \ref{table:tnb2} shows the results 
\david{of our three baselines.} \Gianni{The first criterion we use is the mIoU \cite{jaccard1912distribution}, and the second criterion is the expected calibration error (ECE) \cite{guo2017calibration} that measures how the confidence score predicted by a DNN is related to its accuracy. Finally, we use the AUPR, AU\comEmi{RO}C, and the FPR-95-TPR  defined in \cite{hendrycks2016baseline} that evaluate the quality of a DNN to detect OOD data.}
\david{We can see that} \Gianni{Deep Ensembles outperform other strategies, especially when mixed with Transformers. Yet, MC Dropout seems to have better performance on more complicated sets. Hence MUAD is well suited for quantifying the uncertainty evaluation of different DNNs.}

\begin{table}[t]
\setlength{\abovecaptionskip}{0.cm}
\begin{center}
 \scalebox{0.5}
 {
\begin{tabular}{lcc|c} 
\toprule
\textbf{\Xuanlong{Training set}}&   \textbf{mIoU} $\uparrow$ \\
\midrule
Baseline trained on Cityscapes &  76.84\%  \\
\midrule
Baseline trained on MUAD &  16.71\%  \\
\midrule
Baseline trained on MUAD with histogram eq. &  32.12\%  \\
\midrule
Baseline trained on GTA~\cite{richter2016playing} &  32.85\%  \\
\midrule
Baseline trained on SYNTHIA~\cite{ros2016synthia} &  29.45\%  \\
\bottomrule
\end{tabular}
}
\end{center}
\vspace{-2mm}
\caption{\small \textbf{Comparative results for semantic segmentation simple domain adaptation from MUAD to Cityscapes}. First row is the original baseline, the second row is the performance of the model trained directly on MUAD and the third row is the performance of the model trained on MUAD with histogram matching technique. %
}
\label{table:tnb3_supp}
\vspace{-2mm}
\end{table}

For the second experiment, we train a DeepLabV3+~\cite{chen2018encoder} segmentation network on MUAD and evaluate it on Cityscapes. %
Results reported in Table~\ref{table:tnb3_supp} show that models trained on MUAD images modified with simple histogram matching~\cite{trahanias1992color}\footnote{We use the \texttt{scikit-image}~\cite{van2014scikit} implementation: \url{https://scikit-image.org/docs/dev/api/skimage.exposure.html\#skimage.exposure.match_histograms}} with Cityscapes images achieve the same performance as models trained on the much larger GTA dataset~\cite{richter2016playing}.

\subsection{Monocular depth experiments}

We provide
\david{results}
for monocular depth using NeWCRFs~\cite{yuan2022newcrfs}, which is one of the %
\comEmi{SOTA}
on KITTI dataset~\cite{geiger2012we}.
NeWCRFs %
does not output uncertainty by default. Similarly to \cite{nix1994estimating, kendall2017uncertainties,ilg2018uncertainty}, we modify the DNN to output the parameters of a Gaussian distribution (i.e., the mean and variance). We denote the result as single predictive uncertainty (Single-PU). Based on this modification we train a Deep Ensembles~\cite{lakshminarayanan2017simple} with 3 DNNs. We also provide the results from SLURP~\cite{yu2021slurp}, which needs 2 DNNs to predict the depth and the uncertainty respectively, and MC-Dropout~\cite{gal2016dropout}. For depth evaluation, we use the same %
\ab{metrics} as Eigen et al.~\cite{NIPS2014_7bccfde7} which 
\comEmi{are} %
used in many following works~\cite{lee2019big, yuan2022newcrfs}. For uncertainty quality evaluation, we follow the implementation of Poggi et al.~\cite{poggi2020uncertainty}. More details on implementation and evaluation criteria are provided in the Supplementary Material. 

Table~\ref{table:depth} lists some of the depth and uncertainty results of the above techniques on our dataset \Xuanlong{due to the space limit. We observe that in the presence of OOD, the uncertainty results of Deep Ensembles are comparatively better, while MC-Dropout provides more robust depth estimations under different perturbation.} Additionally, we provide self-supervised monocular depth results using left-right image consistency~\cite{godard2017unsupervised}, along with the full supervised results in Supplementary Material. 
\ab{We also propose a baseline method for depth domain adaptation from MUAD to KITTI and report its performance in Table~\ref{table:kitti_domain_adap}.} 
\Xuanlong{Compared to the direct adaptation from Virtual KITTI 2~\cite{gaidon2016virtual}, which is specifically designed based on the target dataset KITTI~\cite{geiger2012we}, the model trained on MUAD can achieve competitive performance.}

\begin{table}[t]
\centering
\scalebox{0.37}
 {
\begin{tabular}{lccccc|ccccc|cccccccccc} 
\toprule
\multirow{3}{*}{Methods} & \multicolumn{5}{c}{\textbf{normal set}} & \multicolumn{5}{c}{\textbf{low adv. without OOD set}} & \multicolumn{5}{c}{\textbf{high adv. without OOD set}} & \multicolumn{5}{c}{
\comEmi{\textbf{normal set overhead sun}}
} \\
 & \multicolumn{3}{c}{\textbf{Depth results}} & \multicolumn{2}{c}{\textbf{Uncertainty~results}} & \multicolumn{3}{c}{\textbf{\textbf{Depth results}}} & \multicolumn{2}{c}{\textbf{\textbf{Uncertainty~results}}} & \multicolumn{3}{c}{\textbf{\textbf{Depth results}}} & \multicolumn{2}{c}{\textbf{\textbf{Uncertainty~results}}} & \multicolumn{3}{c}{\textbf{\textbf{Depth results}}} & \multicolumn{2}{c}{\textbf{\textbf{Uncertainty~results}}} \\
 & d1 $\uparrow$ & AbsRel $\downarrow$ & RMSE $\downarrow$ & \begin{tabular}[c]{@{}c@{}}AUSE\\RMSE $\downarrow$\end{tabular} & \begin{tabular}[c]{@{}c@{}}AUSE\\Absrel $\downarrow$\end{tabular} & d1 $\uparrow$ & AbsRel $\downarrow$ & RMSE $\downarrow$ & \begin{tabular}[c]{@{}c@{}}AUSE\\RMSE $\downarrow$\end{tabular} & \begin{tabular}[c]{@{}c@{}}AUSE\\Absrel $\downarrow$\end{tabular} & d1 $\uparrow$ & AbsRel $\downarrow$ & RMSE $\downarrow$ & \begin{tabular}[c]{@{}c@{}}AUSE\\RMSE $\downarrow$\end{tabular} & \begin{tabular}[c]{@{}c@{}}AUSE\\Absrel $\downarrow$\end{tabular} & d1 $\uparrow$ & AbsRel $\downarrow$ & RMSE $\downarrow$ & \begin{tabular}[c]{@{}c@{}}AUSE\\RMSE $\downarrow$\end{tabular} & \begin{tabular}[c]{@{}c@{}}AUSE\\Absrel $\downarrow$\end{tabular}\\ 
\midrule
Baseline & \second 0.922 & \second 0.114 & 3.357 & - & - & \second 0.786 & \first \textbf{0.147} & 5.005 & - & - & \second 0.632 & \first\textbf{0.207} & \second 6.989 & - & \multicolumn{1}{c|}{-} & \second 0.951 & \second 0.090 & 3.646 & - & - \\ 
\midrule
Deep Ensembles~\cite{lakshminarayanan2017simple} & \first \textbf{0.929} & \first \textbf{0.111} & \first \textbf{3.199} & \first \textbf{0.291} & \second 0.060 & 0.767 & 0.156 &  4.892 & \first \textbf{0.740} & 0.105 & 0.566 & \second 0.243 & 7.498 & \first \textbf{1.182} & \multicolumn{1}{c|}{\second 0.153} & \first \textbf{0.955} & \first \textbf{0.083} & \second 3.479 & \first \textbf{0.336} & \second 0.055 \\
\midrule
MC Dropout~\cite{gal2016dropout} & 0.919 & 0.119 & \second 3.209 & 0.634 & 0.061 & \first \textbf{0.798} & \second 0.151 & \first \textbf{4.580} & 1.063 & \second 0.098 & \first \textbf{0.657} & \first \textbf{0.207} & \first \textbf{6.278} & \second 1.382 & \first \textbf{0.128} & 0.948 & 0.092 & \first \textbf{3.407} & 0.786 & 0.058 \\ 
\midrule
Single-PU~\cite{kendall2017uncertainties} & 0.905 & 0.132 & 3.230 & \second 0.313 & 0.081 & 0.773 & 0.159 & \second 4.865 & \second 0.789 & 0.112 & 0.571 & 0.248 & 7.680 & 1.740 & \multicolumn{1}{c|}{0.171} & 0.946 & 0.105 & 3.546 & \second 0.358 & 0.079 \\ 
\midrule
SLURP~\cite{yu2021slurp} & \second 0.922 & \second 0.114 & 3.357 & 0.467 & \first \textbf{0.048} & \second 0.786 & \first \textbf{0.147} & 5.005 & 1.167 & \first \textbf{0.090} & \second 0.632 & \first \textbf{0.207} & \second 6.989 & 1.707 & \multicolumn{1}{c|}{\first \textbf{0.128}} & \second 0.951 & \second 0.090 & 3.646 & 0.525 & \first \textbf{0.033} \\ 
\bottomrule
\end{tabular}
}
\vspace{1pt}
\scalebox{0.4}
{
\begin{tabular}{lccccc|ccccc|cccccccccc} 
\cmidrule[\heavyrulewidth]{1-16}
\multirow{3}{*}{Methods} & \multicolumn{5}{c}{\textbf{OOD set}} & \multicolumn{5}{c}{\textbf{low adv. with OOD~set}} & \multicolumn{5}{c}{\textbf{high adv. with OOD set}} & \multicolumn{1}{c}{} &  &  &  &  \\
 & \multicolumn{3}{c}{\textbf{Depth results}} & \multicolumn{2}{c}{\textbf{Uncertainty results}} & \multicolumn{3}{c}{\textbf{\textbf{Depth results}}} & \multicolumn{2}{c}{\textbf{\textbf{Uncertainty results}}} & \multicolumn{3}{c}{\textbf{\textbf{Depth results}}} & \multicolumn{2}{c}{\textbf{\textbf{\textbf{\textbf{Uncertainty results}}}}} &  & \multicolumn{1}{l}{} & \multicolumn{1}{l}{} & \multicolumn{1}{l}{} & \multicolumn{1}{l}{} \\
 & d1 $\uparrow$ & AbsRel $\downarrow$ & RMSE $\downarrow$ & \begin{tabular}[c]{@{}c@{}}AUSE\\RMSE $\downarrow$\end{tabular} & \begin{tabular}[c]{@{}c@{}}AUSE\\Absrel $\downarrow$\end{tabular} & d1 $\uparrow$ & AbsRel $\downarrow$ & RMSE $\downarrow$ & \begin{tabular}[c]{@{}c@{}}AUSE\\RMSE $\downarrow$\end{tabular} & \begin{tabular}[c]{@{}c@{}}AUSE\\Absrel $\downarrow$\end{tabular} & d1 $\uparrow$ & AbsRel $\downarrow$ & RMSE $\downarrow$ & \begin{tabular}[c]{@{}c@{}}AUSE\\RMSE $\downarrow$\end{tabular} & \begin{tabular}[c]{@{}c@{}}AUSE\\Absrel $\downarrow$\end{tabular} &  & \multicolumn{1}{c}{} &  &  & \\
\cmidrule{1-16}
Baseline & \second 0.896 & \second 0.125 & 3.616 & - & - & 0.713 & \second 2.637 & 4.764 & - & - & \second 0.555 & 0.459 & \second 6.916 & - & - &  &  &  &  &  \\ 
\cmidrule{1-16}
Deep Ensembles~\cite{lakshminarayanan2017simple} & \first \textbf{0.903} & \first \textbf{0.114} & \second 3.447 & \first \textbf{0.427} & \second 0.074 & 0.709 & \first \textbf{1.810} & \second 4.707 & \first \textbf{0.692} & \first \textbf{0.129} & 0.521 & \first \textbf{0.331} & 7.411 & \first \textbf{1.072} & \first \textbf{0.151} &  &  &  &  &  \\ 
\cmidrule{1-16}
MC Dropout~\cite{gal2016dropout} & 0.893 & 0.145 & \first \textbf{3.432} & 0.724 & 0.080 & \first \textbf{0.744} & 3.925 & \first \textbf{4.364} & 0.927 & \second 0.206 & \first \textbf{0.610} & 0.545 & \first \textbf{6.176} & \second 1.245 & 0.314 &  &  &  &  &  \\ 
\cmidrule{1-16}
Single-PU~\cite{kendall2017uncertainties} & 0.888 & 0.132 & 3.463 & \second 0.447 & 0.095 & \second 0.714 & 4.349 & 4.716 & \second 0.744 & 0.482 & 0.529 & \second 0.351 & 7.627 & 1.347 & \second 0.156 &  &  &  &  &  \\ 
\cmidrule{1-16}
SLURP~\cite{yu2021slurp} & \second 0.896 & \second 0.125 & 3.616 & 0.721 & \first \textbf{0.068} & 0.713 & \second 2.637 & 4.764 & 1.072 & 0.212 & \second 0.555 & 0.459 & \second 6.916 & 1.564 & \first \textbf{0.151} &  &  &  &  &  \\ 
\cmidrule[\heavyrulewidth]{1-16}
\end{tabular}
}
\vspace{2pt}
\caption{\small \textbf{Comparative results for monocular depth on MUAD}. We use NeWCRFs~\cite{yuan2022newcrfs} as the based DNN for monocular depth task.}
\label{table:depth}
\end{table}

\begin{table*}[t]
\centering
\scalebox{0.45}
{
\begin{tabular}{l|cccccccc} 
\toprule
 & \multicolumn{8}{c}{\textbf{KITTI}} \\
Training set & d1$\uparrow$ & d2$\uparrow$ & d3$\uparrow$ & Abs Rel$\downarrow$ & Sq Rel$\downarrow$ & RMSE$\downarrow$ & RMSE log$\downarrow$ & SILog$\downarrow$ \\ 
\midrule
KITTI~\cite{geiger2012we} & 0.975 & 0.997 & 0.999 & 0.052 & 0.148 & 2.072 & 0.078 & 6.9859 \\ 
\midrule
Virtual KITTI 2~\cite{cabon2020vkitti2} & 0.835 & 0.957 & 0.989 & 0.129 & 0.706 & 4.039 & 0.177 & 15.534 \\ 
\midrule
MUAD & 0.731 & 0.927 & 0.983 & 0.187 & 1.059 & 4.754 & 0.227 & 18.581 \\ 
\bottomrule
\multicolumn{1}{l}{} & \multicolumn{1}{l}{} & \multicolumn{1}{l}{} & \multicolumn{1}{l}{} & \multicolumn{1}{l}{} & \multicolumn{1}{l}{} & \multicolumn{1}{l}{} & \multicolumn{1}{l}{} & \multicolumn{1}{l}{}
\end{tabular}
}
\caption{\small \textbf{Comparative results for monocular depth estimation simple domain adaptation from MUAD to KITTI eigen-split~\cite{NIPS2014_7bccfde7}}. First row is the original baseline, the second and the third rows are the performance of the model trained directly on Virtual KITTI 2~\cite{gaidon2016virtual} and MUAD respectively.%
}
\label{table:kitti_domain_adap}
\end{table*}

\subsection{Object detection experiments}
\begin{table}[t]
\setlength{\abovecaptionskip}{0.cm}
\begin{center}
 \scalebox{0.38}
 {
\begin{tabular}{lccc|ccc|ccc|ccc|ccc|ccc} 
\toprule
\multicolumn{1}{c}{\multirow{2}{*}{Evaluation data}} & \multicolumn{3}{c}{\textbf{normal set}} &\multicolumn{3}{c}{\textbf{low adv. without OOD set}} & \multicolumn{3}{c}{\textbf{high adv. without OOD set}}& \multicolumn{3}{c}{\textbf{OOD set}} & \multicolumn{3}{c}{\textbf{low adv. with OOD set}} & \multicolumn{3}{c}{\textbf{high adv. with OOD set}} \\
 & mAP $\uparrow$ & AP50 $\uparrow$ & PDQ $\uparrow$ & mAP $\uparrow$ & AP50 $\uparrow$ & PDQ $\uparrow$ & mAP $\uparrow$ & AP50 $\uparrow$ & PDQ $\uparrow$ & mAP $\uparrow$ & AP50 $\uparrow$ & PDQ $\uparrow$ & mAP $\uparrow$ & AP50 $\uparrow$ & PDQ $\uparrow$ & mAP $\uparrow$ & AP50 $\uparrow$ & PDQ $\uparrow$\\ 
\midrule
\begin{tabular}[c]{@{}l@{}}Faster R-CNN\\(ResNet101)\end{tabular} & \first \textbf{39.91\% }& \first \textbf{54.91\% }& \first \textbf{16.88\% }&\second 25.00\% &\second 36.89\% &\first \textbf{8.61} &\first \textbf{13.97\%} &\first \textbf{22.01\%} &\first \textbf{0.041} &\first \textbf{35.85\%} & \first \textbf{48.9\%} &\first \textbf{14.33\% } &\first \textbf{24.73\%} &\first  \textbf{35.70\% }&\first \textbf{8.49\%} &\first \textbf{12.41\%} &\first  \textbf{19.66\%} &\first \textbf{3.86\% } \\ 
\midrule
\begin{tabular}[c]{@{}l@{}}Faster R-CNN\\(ResNet50)\end{tabular} &\second 38.43\% &\second 53.13\% &\second 15.02\% &\first \textbf{25.19\%} &\first \textbf{37.38\%} &\second 8.18\% &\second 13.29\% &\second 21.53\% &\second 0.0389\% &\second 34.52\% &\second 47.63\% &\second 12.96\% &\second 23.93\% & \second 34.51\% &\second 7.95\% &\second 12.11\% &\second 19.46\% &\second 3.64\%\\ 
\midrule
\begin{tabular}[c]{@{}l@{}}Gaussian\\YOLOV3\end{tabular}~\cite{choi2019gaussian} & 20.81\% & 32.84\% & 2.22\%  & 8.79\% & 16.40\% & 0.57\% & 3.28\% & 6.30\% & 0.22\% & 17.44\% & 28.16\% & 1.52\% & 10.80\% & 18.71\% & 0.64\%  & 3.21\% & 6.15\% & 0.26\%\\
\bottomrule
\end{tabular}
}
\end{center}
\caption{\small \textbf{Comparative results for object detection on MUAD}. 
The first criteria are the mAP AP50 \cite{lin2014microsoft} related to the accuracy, and the second criterion is the PDQ  \cite{hall2020probabilistic} that measures how well detectors probabilistically localise objects in an image. }
\label{table:Detection}
\vspace{-2pt}
\end{table}

For the object detection task, we trained {a Gaussian YOLOV3~\cite{choi2019gaussian} and} a Faster-RCNN \cite{10.5555/2969239.2969250} on the training data. The Faster R-CNN are trained with ResNet101 and ResNet50 backbones with FPN \cite{8099589}.  All the results are presented on Table \ref{table:Detection}. We can see that {Faster R-CNN's performance drops with} the adversarial conditions, which confirms that considering the adversarial behavior is important when designing algorithms.

\subsection{Discussion}
\comEmi{The experiments show that the best main task contender might
not always be the most suited against different sources of uncertainty, thus it is important to test thoroughly and adapt the processing pipeline to the expected type of perturbations.}
\comEmi{The similar ranking of methods on our synthetic dataset and on real data (see \cite{hendrycks2019anomalyseg,franchi2021robust}) is encouraging as it allows us to generalize the analysis performed on MUAD to actual scenarios. An additional benefit of synthetic datasets is related to the reduced data privacy concerns and regulations that typically affect real world datasets, in particular in urban settings that include pedestrians. All these traits allow for faster validation of new algorithms before their deployment in in real-world settings.}
\ab{Finally, a potential different usage of MUAD concerns unsupervised domain adaptation from synthetic to real domains. Our preliminary results are encouraging.}

\section{Conclusion}
\Gianni{Previous research in deep learning and autonomous cars has established that it is essential to robustify DNNs. In this paper, we present MUAD, a synthetic \comEmi{but highly realistic} dataset 
\comEmi{incorporating multiples sources of}
uncertainties for autonomous driving, that provides insight into the robustness of DNNs for various 
\comEmi{applications}. Based on MUAD, we provide a set of baselines for 
\comEmi{three fundamental}
computer vision tasks. Uncertainty is related to events that occur rarely; synthetic data is very valuable for dealing with infrequent events. We hope that our dataset can improve the reliability of DNNs, especially in autonomous driving scenarios.
We are the first, to our knowledge, to provide a dataset with such 
\comEmi{noise dichotomies present in automotive applications.} 
\comEmi{Our extensive benchmarks show the greater than ever importance of considering uncertainty quantification in addition to accuracy, for decision making systems in sensitive applications.}
}

\clearpage
\section*{Acknowledgement}
We gratefully acknowledge the support of DATAIA Paris-Saclay institute which supported the creation of the dataset (ANR–17–CONV–0003/RD42). This work was performed using HPC resources from GENCI-IDRIS (Grant 2020 - AD011011970) and (Grant 2021 - AD011011970R1) and Saclay-IA computing platform.
\bibliography{egbib}

\clearpage

\title{MUAD: Multiple Uncertainties for Autonomous Driving, a benchmark for multiple uncertainty types and tasks (Supplementary Material)}

\addauthor{Gianni Franchi}{gianni.franchi@ensta-paris.fr}{1$\dagger$}
\addauthor{Xuanlong Yu}{xuanlong.yu@universite-paris-saclay.fr}{1,2$\dagger$}
\addauthor{Andrei Bursuc}{andrei.bursuc@valeo.com}{3}
\addauthor{Angel Tena}{angel.tena@anyverse.ai}{4}
\addauthor{R\'{e}mi Kazmierczak}{remi.kazmierczak@ensta-paris.fr}{1}
\addauthor{S\'{e}verine~Dubuisson}{severine.dubuisson@lis-lab.fr}{5}
\addauthor{Emanuel Aldea}{emanuel.aldea@universite-paris-saclay.fr}{2}
\addauthor{David Filliat}{david.filliat@ensta-paris.fr}{1}
\addinstitution{U2IS, ENSTA Paris, IP Paris}
\addinstitution{SATIE, Paris-Saclay University}
\addinstitution{valeo.ai}
\addinstitution{Anyverse}
\addinstitution{Aix Marseille University}

\begin{small}
\runninghead{MUAD}{Dataset with  Multiple Uncertainties for Autonomous Driving}
\end{small}

\def\eg{\emph{e.g}\bmvaOneDot}
\def\Eg{\emph{E.g}\bmvaOneDot}
\def\etal{\emph{et al}\bmvaOneDot}


\maketitle


\appendix

\setcounter{equation}{4}
\setcounter{figure}{2}
\setcounter{table}{4}

\hspace*{\fill}
\begin{center}
\textbf{\Large MUAD: Multiple Uncertainties for Autonomous Driving, a benchmark for multiple uncertainty types and tasks \\--- Supplementary Material ---}
\end{center}
\hspace*{\fill}

\section{Multiple Uncertainties for Autonomous Driving benchmark (MUAD)}
\label{sec:Datasets}

\subsection{Uncertainty and Deep Learning}

\Gianni{\comEmi{A DNN is a function $f_{\theta}$ parameterized} by a set of parameters $\theta$ that takes input data $x$ and outputs a prediction $y$. \comEmi{The} DNN \comEmi{is} trained on a training dataset composed of a set $\mathcal{D}=\{ x_i,y_i\}_{i=1}^N$, with $N$ being the number of data to optimize the parameters $\theta$ for a task. 
Once the DNN is trained, meaning that the optimization of $\theta$ on   $\mathcal{D}$ \comEmi{is completed},  \comEmi{$f_{\theta}$ may be used for} inference on new data $x^*$.}

\Gianni{Uncertainty on deep learning \comEmi{may arise mainly} from three factors \cite{gawlikowski2021survey}. Firstly it can result from the data acquisition process
\david{which introduces some noise.}
This might be due to the variability in real-world situations. For example, \comEmi{one records training data in} certain weather conditions, \comEmi{which subsequently change} during inferences.
 The measurement systems might also \comEmi{introduce errors such as sensor noise.}
Secondly, \comEmi{uncertainty may} result from the DNN building and training process. DNNs are random functions whose parameters $\theta$ are initialized randomly and whose training procedure relies on stochastic optimization. Therefore, the resulting neural network is a random function that is most of the time \comEmi{related to }a local \comEmi{minimum} of the expected loss function (which we denote as the risk). Hence this source of randomness might cause errors in the training procedure of the DNN.
Thirdly, the last uncertainty factor is related to the DNN's prediction's uncertainty. Uncertainty could come from the lack of knowledge of the DNN and might be caused by unknown test data.}

\Gianni{Based on these factors, we can divide the uncertainty into two kinds: 
the aleatoric uncertainty and the epistemic uncertainty. The aleatoric uncertainty can be subdivided into two kinds: In-domain uncertainty \cite{ashukha2020pitfalls} and Domain-shift uncertainty \cite{ovadia2019can}.
In-domain uncertainty occurs when the test data is sampled from the training distribution and is related to the inability of the deep neural
network to predict a proper confidence score about the quality of its predictions due to a lack of in-domain knowledge.
Domain-shift uncertainty denotes the uncertainty related
to an input drawn from a shifted version of the training
distribution.  Hence, it is caused by the fact the distribution of the training dataset might not encompass enough variability.
These two kinds of uncertainty can be reduced by increasing the
 number of the training dataset.
Epistemic uncertainty denotes the uncertainty when the test data is sampled \comEmi{from} a distribution that is different and far from the training distribution. Epistemic uncertainty can be categorized into two kinds namely \cite{tuna2021exploiting} : approximation uncertainty and model uncertainty. The approximation uncertainty is linked to the fact that we optimize the empirical risk instead of the risk. 
Hence,  the optimal DNN's parameters approximate the optimal DNN's parameters of the true risk function. 
The model uncertainty is linked to the fact that our loss function provides us with a space of solutions that might not include the perfect predictor. For example, the DNN might have  different classes between the training and testing set. In this context 'Out of Distribution' samples refers to anomalies in the test set that are data from classes not present in the training set. }

\section{Extra Monocular depth experiments} 
\subsection{Implementation and criterion}
\Xuanlong{\parag{Implementation.}} \Gianni{We train the NeWCRFs~\cite{yuan2022newcrfs} model using the same hyperparameters and image augmentation parameters used in the official paper for training on KITTI~\cite{geiger2012we}, except that we change the batch size to 4 and randomly crop the input image to 512*1024.
 For the Single-PU~\cite{kendall2017uncertainties} models, we perform a multi task training where we train to predict the depth map with the silog loss function provided in the NeWCRFs~\cite{yuan2022newcrfs} paper, and we minimize the negative Gaussian log-likelihood loss in order to train to predict the variance. 
To train the DNN that will predict the variance, we do not optimize the layers that are used to predict the depth map, as explained in~\cite{Asai_2019_CVPR_Workshops}, as this stabilizes the training. Regarding MC-Dropout~\cite{gal2016dropout}, we let the dropout layers activated during the inference and perform eight forward passes for each input data during inference and average the predictions. We want to point out that we did not add any additional dropout layers to the model to keep the paper's performance. For the SLURP~\cite{yu2021slurp} models, we use the base model as the main task model and train an auxiliary uncertainty estimator. We use the Swin Transformer~\cite{liu2021swin} as used in the base model as an encoder for the auxiliary model and train the auxiliary model for 20 epochs.}
\\\Xuanlong{\parag{Evaluation metrics.}}
\Gianni{To evaluate depth estimations, we use the same metrics as Eigen \etal~\cite{NIPS2014_7bccfde7} which are standard criteria~\cite{lee2019big, zhou_diffnet, yuan2022newcrfs}. For uncertainty quantification
evaluation metrics, we use the criteria implementation of Poggi \etal~\cite{poggi2020uncertainty}: Area Under the Sparsification Error (AUSE) and Area Under the Random Gain (AURG).  The Area Under the Sparsification Error is obtained by calculating the difference between the sparsification curve and the oracle sparsification curve. The sparsification curve is obtained by continuously erasing 1\% pixels according to the predicted uncertainty and calculating the prediction error for the rest pixels. We can also have an oracle sparsification curve by continuously erasing pixels according to their prediction error. The total difference between the two curves is AUSE. We can evaluate the  AUSE for different error metrics such as RMSE, Absrel, and d1, which provide us AUSE-RMSE, AUSE-Absrel, and AUSE-d1.
AURG is achieved by calculating the area between the Sparsification curve and a random curve to measure how good the uncertainty estimator is compared to no modeling cases. Similarly, we can achieve AURG-RMSE, AURG-Absrel, and AURG-d1 using different error metrics.
}
\subsection{Full results on supervised monocular depth estimation}
\Xuanlong{In the main paper, due to the space constrain, we can only provide partial results for depth and uncertainty metrics, we here provide full results from Table~\ref{tab:full_normal} to Table~\ref{tab:full_level2_ood} for different uncertainty quantification solutions introduced in the main paper applied on supervised monocular depth estimation task. Overall, the Deep Ensembles~\cite{lakshminarayanan2017simple} and SLURP~\cite{yu2021slurp} can provide better uncertainty estimations on the test sets without perturbations. When weather perturbations exist, MC-Dropout~\cite{gal2016dropout} and Deep Ensembles~\cite{lakshminarayanan2017simple} perform better on uncertainty quantification. MC-Dropout can also provide better depth estimations than the other solutions under weather perturbations.}

\begin{table}[t]
\centering
\scalebox{0.50}
{
\centering
\begin{tabular}{llllllllll|llllll} 
\toprule
\multicolumn{1}{c}{\multirow{2}{*}{Methods}} & \multicolumn{1}{c}{\multirow{2}{*}{silog$\downarrow$}} & \multicolumn{1}{c}{\multirow{2}{*}{AbsRel$\downarrow$}} & \multicolumn{1}{c}{\multirow{2}{*}{log10$\downarrow$}} & \multicolumn{1}{c}{\multirow{2}{*}{RMSE$\downarrow$}} & \multicolumn{1}{c}{\multirow{2}{*}{SqRel$\downarrow$}} & \multicolumn{1}{c}{\multirow{2}{*}{log\_RMSE}} & \multicolumn{1}{c}{\multirow{2}{*}{d1$\uparrow$}} & \multicolumn{1}{c}{\multirow{2}{*}{d2$\uparrow$}} & \multicolumn{1}{c|}{\multirow{2}{*}{d3$\uparrow$}} & \multicolumn{3}{c}{AUSE$\downarrow$} & \multicolumn{3}{c}{AURG$\uparrow$} \\
\multicolumn{1}{c}{} & \multicolumn{1}{c}{} & \multicolumn{1}{c}{} & \multicolumn{1}{c}{} & \multicolumn{1}{c}{} & \multicolumn{1}{c}{} & \multicolumn{1}{c}{} & \multicolumn{1}{c}{} & \multicolumn{1}{c}{} & \multicolumn{1}{c|}{} & \multicolumn{1}{c}{AbsRel} & \multicolumn{1}{c}{RMSE} & \multicolumn{1}{c}{d1} & \multicolumn{1}{c}{AbsRel} & \multicolumn{1}{c}{RMSE} & \multicolumn{1}{c}{d1} \\ 
\toprule
Baseline & 13.9767 & 0.1143 & 0.0444 & 3.3575 & {\cellcolor[rgb]{0.698,0.698,1}}0.5571 & 0.1443 & 0.9219 & 0.9833 & 0.9933 & \multicolumn{1}{c}{-} & \multicolumn{1}{c}{-} & \multicolumn{1}{c}{-} & \multicolumn{1}{c}{-} & \multicolumn{1}{c}{-} & \multicolumn{1}{c}{-} \\
Deep Ensembles~\cite{lakshminarayanan2017simple} & 13.6691 & {\cellcolor[rgb]{0.698,0.698,1}}0.1110 & {\cellcolor[rgb]{0.698,0.698,1}}0.0419 & {\cellcolor[rgb]{0.698,0.698,1}}3.1994 & 0.6076 & {\cellcolor[rgb]{0.698,0.698,1}}0.1400 & {\cellcolor[rgb]{0.698,0.698,1}}0.9289 & 0.9843 & {\cellcolor[rgb]{0.698,0.698,1}}0.9945 & 0.0604 & {\cellcolor[rgb]{0.698,0.698,1}}0.2906 & {\cellcolor[rgb]{0.698,0.698,1}}0.0431 & 0.0117 & {\cellcolor[rgb]{0.698,0.698,1}}2.4618 & 0.0215 \\
MC Dropout~\cite{gal2016dropout} & {\cellcolor[rgb]{0.698,0.698,1}}13.5602 & 0.1194 & 0.0447 & 3.2090 & 0.6897 & 0.1453 & 0.9193 & {\cellcolor[rgb]{0.698,0.698,1}}0.9847 & 0.9941 & 0.0610 & 0.6339 & 0.0542 & 0.0161 & 2.0846 & 0.0193 \\
Single-PU~\cite{kendall2017uncertainties} & 14.5896 & 0.1324 & 0.0484 & 3.2298 & 0.7738 & 0.1547 & 0.9054 & 0.9803 & 0.9933 & 0.0807 & 0.3131 & 0.0837 & 0.0042 & 2.4194 & -0.0005 \\
SLURP~\cite{yu2021slurp}& 13.9767 & 0.1143 & 0.0444 & 3.3575 & {\cellcolor[rgb]{0.698,0.698,1}}0.5571 & 0.1443 & 0.9219 & 0.9833 & 0.9933 & {\cellcolor[rgb]{0.698,0.698,1}}0.0477 & 0.4672 & 0.0459 & {\cellcolor[rgb]{0.698,0.698,1}}0.0252 & 2.3870 & {\cellcolor[rgb]{0.698,0.698,1}}0.0237\\
\bottomrule
\end{tabular}
}
\caption{{Supervised monocular depth results on \textbf{normal set}.}}
\label{tab:full_normal}
\end{table}

\begin{table}[t]
\centering
\scalebox{0.50}
{
\centering
\begin{tabular}{llllllllll|llllll} 
\toprule
\multicolumn{1}{c}{\multirow{2}{*}{Methods}} & \multicolumn{1}{c}{\multirow{2}{*}{silog$\downarrow$}} & \multicolumn{1}{c}{\multirow{2}{*}{AbsRel$\downarrow$}} & \multicolumn{1}{c}{\multirow{2}{*}{log10$\downarrow$}} & \multicolumn{1}{c}{\multirow{2}{*}{RMSE$\downarrow$}} & \multicolumn{1}{c}{\multirow{2}{*}{SqRel$\downarrow$}} & \multicolumn{1}{c}{\multirow{2}{*}{log\_RMSE}} & \multicolumn{1}{c}{\multirow{2}{*}{d1$\uparrow$}} & \multicolumn{1}{c}{\multirow{2}{*}{d2$\uparrow$}} & \multicolumn{1}{c|}{\multirow{2}{*}{d3$\uparrow$}} & \multicolumn{3}{c}{AUSE$\downarrow$} & \multicolumn{3}{c}{AURG$\uparrow$} \\
\multicolumn{1}{c}{} & \multicolumn{1}{c}{} & \multicolumn{1}{c}{} & \multicolumn{1}{c}{} & \multicolumn{1}{c}{} & \multicolumn{1}{c}{} & \multicolumn{1}{c}{} & \multicolumn{1}{c}{} & \multicolumn{1}{c}{} & \multicolumn{1}{c|}{} & \multicolumn{1}{c}{AbsRel} & \multicolumn{1}{c}{RMSE} & \multicolumn{1}{c}{d1} & \multicolumn{1}{c}{AbsRel} & \multicolumn{1}{c}{RMSE} & \multicolumn{1}{c}{d1} \\ 
\toprule
Baseline & {\cellcolor[rgb]{0.698,0.698,1}}19.8427 & {\cellcolor[rgb]{0.698,0.698,1}}0.1474 & {\cellcolor[rgb]{0.698,0.698,1}}0.0757 & 5.0053 & 0.8301 & {\cellcolor[rgb]{0.698,0.698,1}}0.2397 & 0.7861 & {\cellcolor[rgb]{0.698,0.698,1}}0.9244 & {\cellcolor[rgb]{0.698,0.698,1}}0.9613 & \multicolumn{1}{c}{-} & \multicolumn{1}{c}{-} & \multicolumn{1}{c}{-} & \multicolumn{1}{c}{-} & \multicolumn{1}{c}{-} & \multicolumn{1}{c}{-} \\
Deep Ensembles~\cite{lakshminarayanan2017simple} & 22.7950 & 0.1564 & 0.0850 & 4.8919 & 0.8508 & 0.2759 & 0.7673 & 0.9010 & 0.9419 & 0.1047 & {\cellcolor[rgb]{0.698,0.698,1}}0.7401 & 0.1823 & -0.0103 & {\cellcolor[rgb]{0.698,0.698,1}}3.1624 & 0.0023 \\
MC Dropout~\cite{gal2016dropout} & 21.6959 & 0.1505 & 0.0765 & {\cellcolor[rgb]{0.698,0.698,1}}4.5799 & {\cellcolor[rgb]{0.698,0.698,1}}0.7648 & 0.2459 & {\cellcolor[rgb]{0.698,0.698,1}}0.7980 & 0.9199 & 0.9543 & 0.0980 & 1.0627 & {\cellcolor[rgb]{0.698,0.698,1}}0.1473 & -0.0074 & 2.5851 & {\cellcolor[rgb]{0.698,0.698,1}}0.0182 \\
Single-PU~\cite{kendall2017uncertainties} & 24.2069 & 0.1588 & 0.0849 & 4.8648 & 0.8522 & 0.2800 & 0.7727 & 0.8997 & 0.9417 & 0.1115 & 0.7892 & 0.1863 & -0.0145 & 3.1099 & -0.0025 \\
SLURP~\cite{yu2021slurp}& 19.8429 & {\cellcolor[rgb]{0.698,0.698,1}}0.1474 & {\cellcolor[rgb]{0.698,0.698,1}}0.0757 & 5.0053 & 0.8301 & {\cellcolor[rgb]{0.698,0.698,1}}0.2397 & 0.7861 & {\cellcolor[rgb]{0.698,0.698,1}}0.9244 & {\cellcolor[rgb]{0.698,0.698,1}}0.9613 & {\cellcolor[rgb]{0.698,0.698,1}}0.0898 & 1.1665 & 0.1789 & {\cellcolor[rgb]{0.698,0.698,1}}-0.0040 & 2.8036 & -0.0037

 \\
\bottomrule
\end{tabular}
}
\caption{{Supervised monocular depth results on \textbf{low adv. without OOD set}.}}
\label{tab:full_level1}
\end{table}

\begin{table}[t]
\centering
\scalebox{0.50}
{
\centering
\begin{tabular}{llllllllll|llllll} 
\toprule
\multicolumn{1}{c}{\multirow{2}{*}{Methods}} & \multicolumn{1}{c}{\multirow{2}{*}{silog$\downarrow$}} & \multicolumn{1}{c}{\multirow{2}{*}{AbsRel$\downarrow$}} & \multicolumn{1}{c}{\multirow{2}{*}{log10$\downarrow$}} & \multicolumn{1}{c}{\multirow{2}{*}{RMSE$\downarrow$}} & \multicolumn{1}{c}{\multirow{2}{*}{SqRel$\downarrow$}} & \multicolumn{1}{c}{\multirow{2}{*}{log\_RMSE}} & \multicolumn{1}{c}{\multirow{2}{*}{d1$\uparrow$}} & \multicolumn{1}{c}{\multirow{2}{*}{d2$\uparrow$}} & \multicolumn{1}{c|}{\multirow{2}{*}{d3$\uparrow$}} & \multicolumn{3}{c}{AUSE$\downarrow$} & \multicolumn{3}{c}{AURG$\uparrow$} \\
\multicolumn{1}{c}{} & \multicolumn{1}{c}{} & \multicolumn{1}{c}{} & \multicolumn{1}{c}{} & \multicolumn{1}{c}{} & \multicolumn{1}{c}{} & \multicolumn{1}{c}{} & \multicolumn{1}{c}{} & \multicolumn{1}{c}{} & \multicolumn{1}{c|}{} & \multicolumn{1}{c}{AbsRel} & \multicolumn{1}{c}{RMSE} & \multicolumn{1}{c}{d1} & \multicolumn{1}{c}{AbsRel} & \multicolumn{1}{c}{RMSE} & \multicolumn{1}{c}{d1} \\ 
\toprule
Baseline & {\cellcolor[rgb]{0.698,0.698,1}}27.2917 & {\cellcolor[rgb]{0.698,0.698,1}}0.2072 & 0.1148 & 6.9890 & 1.5990 & {\cellcolor[rgb]{0.698,0.698,1}}0.3603 & 0.6316 & 0.8275 & {\cellcolor[rgb]{0.698,0.698,1}}0.9028 & \multicolumn{1}{c}{-} & \multicolumn{1}{c}{-} & \multicolumn{1}{c}{-} & \multicolumn{1}{c}{-} & \multicolumn{1}{c}{-} & \multicolumn{1}{c}{-} \\
Deep Ensembles~\cite{lakshminarayanan2017simple} & 34.7624 & 0.2429 & 0.1478 & 7.4977 & 1.9794 & 0.4674 & 0.5657 & 0.7643 & 0.8507 & 0.1529 & {\cellcolor[rgb]{0.698,0.698,1}}1.1824 & 0.3031 & -0.0117 & {\cellcolor[rgb]{0.698,0.698,1}}4.6140 & -0.0044 \\
MC Dropout~\cite{gal2016dropout} & 30.5442 & 0.2073 & {\cellcolor[rgb]{0.698,0.698,1}}0.1142 & {\cellcolor[rgb]{0.698,0.698,1}}6.2782 & {\cellcolor[rgb]{0.698,0.698,1}}1.3762 & 0.3652 & {\cellcolor[rgb]{0.698,0.698,1}}0.6567 & {\cellcolor[rgb]{0.698,0.698,1}}0.8292 & 0.8992 & {\cellcolor[rgb]{0.698,0.698,1}}0.1277 & 1.3819 & {\cellcolor[rgb]{0.698,0.698,1}}0.2169 & {\cellcolor[rgb]{0.698,0.698,1}}-0.0055 & 3.5187 & {\cellcolor[rgb]{0.698,0.698,1}}0.0393 \\
Single-PU~\cite{kendall2017uncertainties} & 41.9847 & 0.2480 & 0.1588 & 7.6797 & 2.1362 & 0.5295 & 0.5708 & 0.7586 & 0.8435 & 0.1706 & 1.7402 & 0.3318 & -0.0220 & 4.2634 & -0.0322 \\
SLURP~\cite{yu2021slurp}& {\cellcolor[rgb]{0.698,0.698,1}}27.2917 & {\cellcolor[rgb]{0.698,0.698,1}}0.2072 & 0.1148 & 6.9890 & 1.5990 & {\cellcolor[rgb]{0.698,0.698,1}}0.3603 & 0.6316 & 0.8275 & {\cellcolor[rgb]{0.698,0.698,1}}0.9028 & 0.1281 & 1.7066 & 0.2740 & -0.0100 & 3.7188 & -0.0024

\\
\bottomrule
\end{tabular}
}
\caption{{Supervised monocular depth results on \textbf{high adv. without OOD set}}.}
\label{tab:full_level2}
\end{table}

\begin{table}[t]
\centering
\scalebox{0.50}
{
\centering
\begin{tabular}{llllllllll|llllll} 
\toprule
\multicolumn{1}{c}{\multirow{2}{*}{Methods}} & \multicolumn{1}{c}{\multirow{2}{*}{silog$\downarrow$}} & \multicolumn{1}{c}{\multirow{2}{*}{AbsRel$\downarrow$}} & \multicolumn{1}{c}{\multirow{2}{*}{log10$\downarrow$}} & \multicolumn{1}{c}{\multirow{2}{*}{RMSE$\downarrow$}} & \multicolumn{1}{c}{\multirow{2}{*}{SqRel$\downarrow$}} & \multicolumn{1}{c}{\multirow{2}{*}{log\_RMSE}} & \multicolumn{1}{c}{\multirow{2}{*}{d1$\uparrow$}} & \multicolumn{1}{c}{\multirow{2}{*}{d2$\uparrow$}} & \multicolumn{1}{c|}{\multirow{2}{*}{d3$\uparrow$}} & \multicolumn{3}{c}{AUSE$\downarrow$} & \multicolumn{3}{c}{AURG$\uparrow$} \\
\multicolumn{1}{c}{} & \multicolumn{1}{c}{} & \multicolumn{1}{c}{} & \multicolumn{1}{c}{} & \multicolumn{1}{c}{} & \multicolumn{1}{c}{} & \multicolumn{1}{c}{} & \multicolumn{1}{c}{} & \multicolumn{1}{c}{} & \multicolumn{1}{c|}{} & \multicolumn{1}{c}{AbsRel} & \multicolumn{1}{c}{RMSE} & \multicolumn{1}{c}{d1} & \multicolumn{1}{c}{AbsRel} & \multicolumn{1}{c}{RMSE} & \multicolumn{1}{c}{d1} \\ 
\toprule
Baseline & 12.4227 & 0.0895 & 0.0387 & 3.6461 & 0.4083 & 0.1257 & 0.9513 & {\cellcolor[rgb]{0.698,0.698,1}}0.9909 & {\cellcolor[rgb]{0.698,0.698,1}}0.9969 & \multicolumn{1}{c}{-} & \multicolumn{1}{c}{-} & \multicolumn{1}{c}{-} & \multicolumn{1}{c}{-} & \multicolumn{1}{c}{-} & \multicolumn{1}{c}{-} \\
Deep Ensembles~\cite{lakshminarayanan2017simple} & {\cellcolor[rgb]{0.698,0.698,1}}11.7212 & {\cellcolor[rgb]{0.698,0.698,1}}0.0829 & {\cellcolor[rgb]{0.698,0.698,1}}0.0351 & 3.4788 & {\cellcolor[rgb]{0.698,0.698,1}}0.3867 & {\cellcolor[rgb]{0.698,0.698,1}}0.1188 & {\cellcolor[rgb]{0.698,0.698,1}}0.9553 & 0.9903 & 0.9967 & 0.0553 & {\cellcolor[rgb]{0.698,0.698,1}}0.3363 & {\cellcolor[rgb]{0.698,0.698,1}}0.0098 & -0.0041 & {\cellcolor[rgb]{0.698,0.698,1}}2.6248 & 0.0336 \\
MC Dropout~\cite{gal2016dropout} & 12.0129 & 0.0915 & 0.0389 & {\cellcolor[rgb]{0.698,0.698,1}}3.4074 & 0.3888 & 0.1263 & 0.9475 & 0.9902 & {\cellcolor[rgb]{0.698,0.698,1}}0.9969 & 0.0576 & 0.7856 & 0.0308 & -0.0019 & 2.0452 & 0.0199 \\
Single-PU~\cite{kendall2017uncertainties} & 12.4754 & 0.1052 & 0.0437 & 3.5463 & 0.4210 & 0.1344 & 0.9461 & 0.9895 & 0.9966 & 0.0788 & 0.3576 & 0.0308 & -0.0189 & 2.5430 & 0.0212 \\
SLURP~\cite{yu2021slurp}& 12.4227 & 0.0895 & 0.0387 & 3.6461 & 0.4083 & 0.1257 & 0.9513 & {\cellcolor[rgb]{0.698,0.698,1}}0.9909 & {\cellcolor[rgb]{0.698,0.698,1}}0.9969 & {\cellcolor[rgb]{0.698,0.698,1}}0.0328 & 0.5248 & 0.0100 & {\cellcolor[rgb]{0.698,0.698,1}}0.0222 & 2.5207 & {\cellcolor[rgb]{0.698,0.698,1}}0.0373
\\
\bottomrule
\end{tabular}
}
\caption{{Supervised monocular depth results on \textbf{normal test set with Overhead Sun}}.}
\label{tab:full_noon}
\end{table}

\begin{table}[t]
\centering
\scalebox{0.50}
{
\centering
\begin{tabular}{llllllllll|llllll} 
\toprule
\multicolumn{1}{c}{\multirow{2}{*}{Methods}} & \multicolumn{1}{c}{\multirow{2}{*}{silog$\downarrow$}} & \multicolumn{1}{c}{\multirow{2}{*}{AbsRel$\downarrow$}} & \multicolumn{1}{c}{\multirow{2}{*}{log10$\downarrow$}} & \multicolumn{1}{c}{\multirow{2}{*}{RMSE$\downarrow$}} & \multicolumn{1}{c}{\multirow{2}{*}{SqRel$\downarrow$}} & \multicolumn{1}{c}{\multirow{2}{*}{log\_RMSE}} & \multicolumn{1}{c}{\multirow{2}{*}{d1$\uparrow$}} & \multicolumn{1}{c}{\multirow{2}{*}{d2$\uparrow$}} & \multicolumn{1}{c|}{\multirow{2}{*}{d3$\uparrow$}} & \multicolumn{3}{c}{AUSE$\downarrow$} & \multicolumn{3}{c}{AURG$\uparrow$} \\
\multicolumn{1}{c}{} & \multicolumn{1}{c}{} & \multicolumn{1}{c}{} & \multicolumn{1}{c}{} & \multicolumn{1}{c}{} & \multicolumn{1}{c}{} & \multicolumn{1}{c}{} & \multicolumn{1}{c}{} & \multicolumn{1}{c}{} & \multicolumn{1}{c|}{} & \multicolumn{1}{c}{AbsRel} & \multicolumn{1}{c}{RMSE} & \multicolumn{1}{c}{d1} & \multicolumn{1}{c}{AbsRel} & \multicolumn{1}{c}{RMSE} & \multicolumn{1}{c}{d1} \\ 
\toprule
Baseline & 16.4332 & 0.1250 & 0.0525 & 3.6157 & 0.5875 & 0.1747 & 0.8956 & 0.9602 & 0.9783 & \multicolumn{1}{c}{-} & \multicolumn{1}{c}{-} & \multicolumn{1}{c}{-} & \multicolumn{1}{c}{-} & \multicolumn{1}{c}{-} & \multicolumn{1}{c}{-} \\
Deep Ensembles~\cite{lakshminarayanan2017simple} & 16.3795 & {\cellcolor[rgb]{0.698,0.698,1}}0.1142 & {\cellcolor[rgb]{0.698,0.698,1}}0.0503 & 3.4465 & {\cellcolor[rgb]{0.698,0.698,1}}0.4812 & {\cellcolor[rgb]{0.698,0.698,1}}0.1724 & {\cellcolor[rgb]{0.698,0.698,1}}0.9027 & 0.9600 & 0.9777 & 0.0739 & {\cellcolor[rgb]{0.698,0.698,1}}0.4268 & {\cellcolor[rgb]{0.698,0.698,1}}0.0563 & -0.0016 & {\cellcolor[rgb]{0.698,0.698,1}}2.4750 & {\cellcolor[rgb]{0.698,0.698,1}}0.0296 \\
MC Dropout~\cite{gal2016dropout} & {\cellcolor[rgb]{0.698,0.698,1}}16.1976 & 0.1277 & 0.0525 & {\cellcolor[rgb]{0.698,0.698,1}}3.4437 & 0.5923 & 0.1744 & 0.8934 & {\cellcolor[rgb]{0.698,0.698,1}}0.9620 & {\cellcolor[rgb]{0.698,0.698,1}}0.9799 & 0.0720 & 0.7253 & 0.0649 & 0.0104 & 2.1331 & 0.0292 \\
Single-PU~\cite{kendall2017uncertainties} & 17.1019 & 0.1319 & 0.0561 & 3.4628 & 0.5126 & 0.1833 & 0.8884 & 0.9580 & 0.9777 & 0.0948 & 0.4474 & 0.0872 & -0.0135 & 2.4091 & 0.0103 \\
SLURP~\cite{yu2021slurp}& 16.4332 & 0.1250 & 0.0525 & 3.6157 & 0.5875 & 0.1747 & 0.8956 & 0.9602 & 0.9783 & {\cellcolor[rgb]{0.698,0.698,1}}0.0681 & 0.7208 & 0.0852 & {\cellcolor[rgb]{0.698,0.698,1}}0.0121 & 2.2899 & 0.0054
\\
\bottomrule
\end{tabular}
}
\caption{{Supervised monocular depth results on \textbf{OOD set}}.}
\label{tab:full_ood}
\end{table}

\begin{table}[t]
\centering
\scalebox{0.50}
{
\centering
\begin{tabular}{llllllllll|llllll} 
\toprule
\multicolumn{1}{c}{\multirow{2}{*}{Methods}} & \multicolumn{1}{c}{\multirow{2}{*}{silog$\downarrow$}} & \multicolumn{1}{c}{\multirow{2}{*}{AbsRel$\downarrow$}} & \multicolumn{1}{c}{\multirow{2}{*}{log10$\downarrow$}} & \multicolumn{1}{c}{\multirow{2}{*}{RMSE$\downarrow$}} & \multicolumn{1}{c}{\multirow{2}{*}{SqRel$\downarrow$}} & \multicolumn{1}{c}{\multirow{2}{*}{log\_RMSE}} & \multicolumn{1}{c}{\multirow{2}{*}{d1$\uparrow$}} & \multicolumn{1}{c}{\multirow{2}{*}{d2$\uparrow$}} & \multicolumn{1}{c|}{\multirow{2}{*}{d3$\uparrow$}} & \multicolumn{3}{c}{AUSE$\downarrow$} & \multicolumn{3}{c}{AURG$\uparrow$} \\
\multicolumn{1}{c}{} & \multicolumn{1}{c}{} & \multicolumn{1}{c}{} & \multicolumn{1}{c}{} & \multicolumn{1}{c}{} & \multicolumn{1}{c}{} & \multicolumn{1}{c}{} & \multicolumn{1}{c}{} & \multicolumn{1}{c}{} & \multicolumn{1}{c|}{} & \multicolumn{1}{c}{AbsRel} & \multicolumn{1}{c}{RMSE} & \multicolumn{1}{c}{d1} & \multicolumn{1}{c}{AbsRel} & \multicolumn{1}{c}{RMSE} & \multicolumn{1}{c}{d1} \\ 
\toprule
Baseline & {\cellcolor[rgb]{0.698,0.698,1}}24.2098 & 2.6367 & 0.0980 & 4.7962 & 10.3942 & 0.3066 & 0.7134 & 0.8775 & 0.9280 & \multicolumn{1}{c}{-} & \multicolumn{1}{c}{-} & \multicolumn{1}{c}{-} & \multicolumn{1}{c}{-} & \multicolumn{1}{c}{-} & \multicolumn{1}{c}{-} \\
Deep Ensembles~\cite{lakshminarayanan2017simple} & 25.9658 & {\cellcolor[rgb]{0.698,0.698,1}}1.8097 & 0.1009 & 4.7072 & {\cellcolor[rgb]{0.698,0.698,1}}5.1183 & 0.3237 & 0.7091 & 0.8652 & 0.9174 & {\cellcolor[rgb]{0.698,0.698,1}}0.1292 & {\cellcolor[rgb]{0.698,0.698,1}}0.6917 & 0.2091 & {\cellcolor[rgb]{0.698,0.698,1}}0.1164 & {\cellcolor[rgb]{0.698,0.698,1}}3.1474 & 0.0067 \\
MC Dropout~\cite{gal2016dropout} & 25.3372 & 3.9252 & {\cellcolor[rgb]{0.698,0.698,1}}0.0924 & {\cellcolor[rgb]{0.698,0.698,1}}4.3635 & 22.9193 & {\cellcolor[rgb]{0.698,0.698,1}}0.2971 & {\cellcolor[rgb]{0.698,0.698,1}}0.7437 & {\cellcolor[rgb]{0.698,0.698,1}}0.8829 & {\cellcolor[rgb]{0.698,0.698,1}}0.9287 & 0.2062 & 0.9267 & {\cellcolor[rgb]{0.698,0.698,1}}0.1843 & 0.0598 & 2.6365 & {\cellcolor[rgb]{0.698,0.698,1}}0.0125 \\
Single-PU~\cite{kendall2017uncertainties} & 27.3008 & 4.3492 & 0.1009 & 4.7161 & 28.5999 & 0.3284 & 0.7140 & 0.8638 & 0.9174 & 0.4815 & 0.7444 & 0.2104 & -0.0210 & 3.1238 & 0.0039 \\
SLURP~\cite{yu2021slurp}& {\cellcolor[rgb]{0.698,0.698,1}}24.2098 & 2.6366 & 0.0980 & 4.7962 & 10.3930 & 0.3066 & 0.7134 & 0.8775 & 0.9280 & 0.2116 & 1.0715 & 0.2229 & 0.0682 & 2.8043 & -0.0116

\\
\bottomrule
\end{tabular}
}
\caption{{Supervised monocular depth results on \textbf{low adv. with OOD set}}.}
\label{tab:full_level1_ood}
\end{table}

\begin{table}[t]
\centering
\scalebox{0.50}
{
\centering
\begin{tabular}{llllllllll|llllll} 
\toprule
\multicolumn{1}{c}{\multirow{2}{*}{Methods}} & \multicolumn{1}{c}{\multirow{2}{*}{silog$\downarrow$}} & \multicolumn{1}{c}{\multirow{2}{*}{AbsRel$\downarrow$}} & \multicolumn{1}{c}{\multirow{2}{*}{log10$\downarrow$}} & \multicolumn{1}{c}{\multirow{2}{*}{RMSE$\downarrow$}} & \multicolumn{1}{c}{\multirow{2}{*}{SqRel$\downarrow$}} & \multicolumn{1}{c}{\multirow{2}{*}{log\_RMSE}} & \multicolumn{1}{c}{\multirow{2}{*}{d1$\uparrow$}} & \multicolumn{1}{c}{\multirow{2}{*}{d2$\uparrow$}} & \multicolumn{1}{c|}{\multirow{2}{*}{d3$\uparrow$}} & \multicolumn{3}{c}{AUSE$\downarrow$} & \multicolumn{3}{c}{AURG$\uparrow$} \\
\multicolumn{1}{c}{} & \multicolumn{1}{c}{} & \multicolumn{1}{c}{} & \multicolumn{1}{c}{} & \multicolumn{1}{c}{} & \multicolumn{1}{c}{} & \multicolumn{1}{c}{} & \multicolumn{1}{c}{} & \multicolumn{1}{c}{} & \multicolumn{1}{c|}{} & \multicolumn{1}{c}{AbsRel} & \multicolumn{1}{c}{RMSE} & \multicolumn{1}{c}{d1} & \multicolumn{1}{c}{AbsRel} & \multicolumn{1}{c}{RMSE} & \multicolumn{1}{c}{d1} \\ 
\toprule
Baseline & {\cellcolor[rgb]{0.698,0.698,1}}32.1516 & 0.4588 & 0.1448 & 6.9160 & 10.0794 & 0.4422 & 0.5549 & 0.7727 & 0.8587 & \multicolumn{1}{c}{-} & \multicolumn{1}{c}{-} & \multicolumn{1}{c}{-} & \multicolumn{1}{c}{-} & \multicolumn{1}{c}{-} & \multicolumn{1}{c}{-} \\
Deep Ensembles~\cite{lakshminarayanan2017simple} & 37.4423 & {\cellcolor[rgb]{0.698,0.698,1}}0.3308 & 0.1672 & 7.4105 & {\cellcolor[rgb]{0.698,0.698,1}}2.7108 & 0.5183 & 0.5209 & 0.7277 & 0.8179 & {\cellcolor[rgb]{0.698,0.698,1}}0.1509 & {\cellcolor[rgb]{0.698,0.698,1}}1.0724 & 0.2720 & 0.0347 & {\cellcolor[rgb]{0.698,0.698,1}}4.8398 & 0.0285 \\
MC Dropout~\cite{gal2016dropout} & 34.0965 & 0.5448 & {\cellcolor[rgb]{0.698,0.698,1}}0.1351 & {\cellcolor[rgb]{0.698,0.698,1}}6.1764 & 14.0074 & {\cellcolor[rgb]{0.698,0.698,1}}0.4229 & {\cellcolor[rgb]{0.698,0.698,1}}0.6096 & {\cellcolor[rgb]{0.698,0.698,1}}0.7933 & {\cellcolor[rgb]{0.698,0.698,1}}0.8672 & 0.3137 & 1.2454 & {\cellcolor[rgb]{0.698,0.698,1}}0.2394 & 0.0811 & 3.7196 & {\cellcolor[rgb]{0.698,0.698,1}}0.0288 \\
Single-PU~\cite{kendall2017uncertainties} & 42.7338 & 0.3513 & 0.1735 & 7.6272 & 5.0461 & 0.5606 & 0.5289 & 0.7224 & 0.8106 & 0.1556 & 1.3474 & 0.2768 & 0.0611 & 4.7969 & 0.0232 \\
SLURP~\cite{yu2021slurp}& {\cellcolor[rgb]{0.698,0.698,1}}32.1516 & 0.4588 & 0.1448 & 6.9160 & 10.0794 & 0.4422 & 0.5549 & 0.7727 & 0.8587 & 0.1514 & 1.5640 & 0.2737 & {\cellcolor[rgb]{0.698,0.698,1}}0.1437 & 3.9450 & 0.0134

\\
\bottomrule
\end{tabular}
}
\caption{{Supervised monocular depth results on \textbf{high adv. with OOD set}}.}
\label{tab:full_level2_ood}
\end{table}

\subsection{Self-supervised monocular depth estimation}
\Xuanlong{In this section, we provide the self-supervised monocular depth results for MUAD. In order to provide a wider variety of urban scenarios, there are no consecutive frames in MUAD, but still provides pictures taken by the left and right cameras. We provide self-supervised monocular depth results on MUAD in Table~\ref{tab:unsupervised_depth} using DIFFNet~\cite{zhou_diffnet} and left-right consistency~\cite{godard2017unsupervised} strategy. DIFFNet is one of the SOTA on KITTI outdoor dataset~\cite{geiger2012we}. We train a DIFFNet model with 12 images as the batch size, randomly crop the image to 512*1024, and train 20 epochs in total.}

\Gianni{We observe that OOD objects have less impact on the results of monocular depth estimation in the Self-supervised monocular depth. According to~\cite{dijk2019neural}, monocular depth estimation based on left-right coherence is sensitive to illumination conditions, particularly to object shadows. However, our results on the \textit{Normal set} and \textit{Overhead sun set} do not seem to confirm this point. We believe that DNNs learn depth without necessarily paying much attention to shadows; hence they have no impact on the performance of the self-supervised monocular depth model.}

\begin{table}[t]
\centering
\scalebox{0.70}
{
\begin{tabular}{lcccccccc} 
\toprule
Evaluation sets & AbsRel $\downarrow$ & log10 $\downarrow$ & RMSE $\downarrow$ & SqRel $\downarrow$ & log\_RMSE $\downarrow$ & d1 $\uparrow$ & d2 $\uparrow$ & d3 $\uparrow$ \\ 
\toprule
Normal & 0.365 & 0.111 & 5.646 & 2.234 & 0.350 & 0.638 & 0.874 & 0.919 \\
Overhead sun & 0.174 & 0.079 & 5.875 & 1.426 & 0.249 & 0.693 & 0.953 & 0.978 \\
low adv. without OOD & 0.312 & 0.185 & 10.472 & 3.951 & 0.586 & 0.442 & 0.716 & 0.824 \\
high adv. without OOD & 0.510 & 0.432 & 15.578 & 8.513 & 1.194 & 0.227 & 0.417 & 0.531 \\
OOD & 0.312 & 0.101 & 6.170 & 2.663 & 0.331 & 0.648 & 0.899 & 0.941 \\
low adv. with OOD & 1.462 & 0.192 & 9.356 & 6.054 & 0.601 & 0.431 & 0.697 & 0.807 \\
high adv. with OOD & 1.141 & 0.415 & 14.415 & 25.281 & 1.194 & 0.236 & 0.426 & 0.543\\
\bottomrule
\end{tabular}
}
\vspace{2mm}
\caption{{Self-supervised monocular depth results on all test sets given by DIFFNet~\cite{zhou_diffnet}.}}
\label{tab:unsupervised_depth}
\end{table}

\clearpage

\end{document}